\documentclass{article}

\usepackage[table]{xcolor}
\usepackage[final]{corl_2026} % Uncomment for the camera-ready ``final'' version.
\makeatletter
\renewcommand{\@conferencelocation}{Austin, Texas, USA}
\makeatother
\usepackage{amsmath,amssymb,booktabs}
\usepackage{algorithm,algpseudocode}

\definecolor{notationpurple}{RGB}{128,0,128}

\definecolor{finalteal}{RGB}{0,115,115}

\definecolor{carlaedit}{RGB}{150,35,65}

\definecolor{promptcontract}{RGB}{82,92,105}
\definecolor{promptactor}{RGB}{49,108,185}
\definecolor{promptcritic}{RGB}{40,135,98}
\definecolor{promptrepair}{RGB}{184,120,36}
\definecolor{promptworld}{RGB}{113,73,156}
\usepackage{enumitem}
\usepackage{threeparttable}
\usepackage{booktabs}
\usepackage{array}
\usepackage{marvosym}
\usepackage[frozencache,cachedir=minted-cache]{minted}
\usepackage{tcolorbox}
\tcbuselibrary{minted,skins, breakable}

\usepackage{tabularx}
\usepackage{graphicx}
\usepackage{capt-of}
\usepackage{wrapfig}
\usepackage{titletoc}     

\definecolor{mylightblue}{RGB}{100,149,237}
\tcbset{
  enhanced, % Use enhanced skin
  colback=white!200!black, % Background color
  colframe=mylightblue, % Border color
  colbacktitle=mylightblue, % Title background color
  title filled, % Fill the title background
  coltitle=white, % Title text color
  fonttitle=\bfseries, % Title font
  arc=3mm, % Border corner radius
  outer arc=3mm, % Outer border corner radius
  boxrule=0.5mm, % Border width
  toprule=0.5mm, % Top border width
  bottomrule=0.5mm, % Bottom border width
  titlerule=0.5mm, % Border width below the title
  drop fuzzy shadow, % Shadow effect
}
\tcbset{
  promptbox/.style={
    enhanced,
    breakable,
    width=0.98\linewidth,
    boxrule=0.45mm,
    arc=2.2mm,
    outer arc=2.2mm,
    left=2.2mm,
    right=2.2mm,
    top=1.2mm,
    bottom=1.2mm,
    before skip=7pt,
    after skip=8pt,
    fonttitle=\bfseries\sffamily,
    coltitle=white,
    title filled,
    titlerule=0.35mm,
    drop fuzzy shadow
  }
}
\newcommand{\method}{\textsc{EvoDrive}}
\newcommand{\egopolicy}{\pi_e}
\newcommand{\seedset}{\mathcal{S}}
\newcommand{\simulator}{\mathcal{M}}
\newcommand{\operators}{\mathcal{O}_{\mathcal{M}}}
\newcommand{\archive}{\mathcal{P}}
\newcommand{\history}{\mathcal{H}}

\newcommand{\validator}{V}
\newcommand{\compiler}{\Omega}
\newcommand{\world}{W_{\phi}}
\newcommand{\algcompact}{%
  \footnotesize
  \renewcommand{\algorithmicindent}{1.05em}%
}

\title{EvoDrive: Pareto Evolution for Safety-Critical Autonomous Driving via Self-Improving LLM Agents}
\newcommand{\equalcontrib}{\ensuremath{\dagger}}
\newcommand{\corresponding}{\texorpdfstring{\Letter}{*}}

\author{
  {Tong Nie,\textsuperscript{1,2,\equalcontrib}
  Yuewen Mei,\textsuperscript{2,\equalcontrib}  
  Yihong Tang,\textsuperscript{3,4} 
  Junlin He,\textsuperscript{1}
  Jie Deng,\textsuperscript{1} 
  Jian Sun,\textsuperscript{2,\corresponding}
  Wei Ma\textsuperscript{1,\corresponding}}
  \\[2mm]
  {\textsuperscript{1}The Hong Kong Polytechnic University,\,
  \textsuperscript{2}Tongji University, \,}
  \\
  {\textsuperscript{3}McGill University,\,
  \textsuperscript{4}Mila-Quebec AI Institute}
  \\[2mm]
  {\texttt{\href{mailto:tong.nie@connect.polyu.hk}{tong.nie@connect.polyu.hk}}
  \;
  \texttt{\href{mailto:meiyuewen@tongji.edu.cn}{meiyuewen@tongji.edu.cn}}
  \;
  \texttt{\href{mailto:wei.w.ma@polyu.edu.hk}{wei.w.ma@polyu.edu.hk}}}
  \\[1mm]
  {{\normalfont\footnotesize
  \textsuperscript{\equalcontrib} Equal contribution.
  \quad
  \textsuperscript{\corresponding} Corresponding authors.}}
  \\[1mm]
  {\url{https://tongnie.github.io/EvoDrive/}}
}

\hypersetup{
  pdftitle={EvoDrive: Pareto Evolution for Safety-Critical Autonomous Driving via Self-Improving LLM Agents},
  pdfauthor={Tong Nie, Yuewen Mei, Yihong Tang, Junlin He, Jie Deng, Jian Sun, Wei Ma}
}

\begin{document}
\maketitle

%===============================================================================

\begin{abstract}
    Generating safety-critical scenarios is essential for validating and improving autonomous driving systems, yet it inherently requires maximizing adversariality to expose failures while preserving realism. Existing methods usually manage this trade-off with handcrafted heuristics, confining generation to known priors and overlooking underexplored patterns. While recent open-ended agentic evolution can push this limit, unconstrained general agents lack strict simulator grounding and tend to collapse the multi-objective tension into single-scalar maximization. Here we present \method{}, the first automated, LLM-based agentic evolution framework for multi-objective scenario generation. \method{} employs a simulator-grounded actor-critic architecture where a memory-driven actor iteratively proposes improvements to the generators and critics filter out implausible candidates, and a self-evolving world evaluator routes promising proposals to optimize simulation budgets. \method{} further maintains a Pareto archive of evaluated candidates to preserve diverse attack-realism trade-offs and guide future evolution via simulation feedback. Benchmark results on MetaDrive and CARLA show that \method{} not only significantly expands the Pareto frontier across various generators, but also produces valuable scenarios for policy training. 
\end{abstract}


% Main-paper float spacing; the original settings resume before references.
\begingroup
\setlength{\textfloatsep}{10pt plus 2pt minus 2pt}
\setlength{\intextsep}{8pt plus 2pt minus 2pt}
\setlength{\floatsep}{8pt plus 2pt minus 2pt}
\setlength{\dbltextfloatsep}{10pt plus 2pt minus 2pt}
\setlength{\dblfloatsep}{8pt plus 2pt minus 2pt}
\setlength{\parskip}{2pt}
\setlength{\abovedisplayskip}{4pt plus 1pt minus 2pt}
\setlength{\belowdisplayskip}{4pt plus 1pt minus 2pt}
\setlength{\belowdisplayshortskip}{2.5pt plus 1pt minus 1pt}

%===============================================================================

% 这里最应该突出的 gap 是：现有方法依赖人工设计，因此只能在已知模式附近搜索；agentic evolution 有潜力扩大搜索，但如果没有自动驾驶约束，就会产生不真实、不可验证的候选。 ！！！现有启发式方法太受限，而通用的 agentic 框架又太“放飞自我”（缺乏约束）且无法处理多目标。！！！

% 传统方法的问题不是“效果不好”这么泛，而是它们通常只能在 designer-specified perturbation templates / validity filters / scalarized objectives 里面调参，难以发现新的 adversarial mechanisms，也容易把本来应当维护的 Pareto structure 压缩成一个单点折中。

% 通用 agentic evolution 的问题则不同：它能扩大生成器设计空间，但如果没有自动驾驶领域的 generator contracts、simulator-validity checks、realism critics、protected evaluator boundary 和 Pareto acceptance，它会产生不可执行、不真实、不可审计，甚至污染评测边界的候选。这个表述和你当前 EvoDrive 的系统定位一致：你的框架强调不是优化单一加权分数，而是维护 attack-realism Pareto frontier，并通过 role-specialized agents、lineage constraints 和 structured memory 生成新的 generator variants；同时它有 adapter contract、candidate validation、bounded patch 和 deterministic archive judge 等边界。

% \vspace{-5pt}
\section{Introduction}

\vspace{-4pt}
Safety validation for autonomous driving requires testing driving systems under rare but critical scenarios before deployment. Since such scenarios are rare in the real world, closed-loop simulation with scenario generation has become a practical tool for creating controlled stress cases \cite{feng2021nade,feng2023d2rl,xu2022safebench}. However, generating useful adversarial scenarios is an inherently coupled, multi-objective problem \cite{nie2025steerable}. A generator must maximize adversariality to expose policy weaknesses while strictly preserving physical and semantic realism to ensure the resulting interactions are practically meaningful \cite{feng2026breaking, rempe2022generating}.

% Existing safety-critical scenario generation methods attempt to address this trade-off by assigning scalar weights \cite{nie2025steerable}, adding rule-based filters \cite{chen2024frea}, or restricting the search to learned traffic priors \cite{xu2025diffscene, liu2026adv}. While these mechanisms can be effective within specific pipelines, relying on predefined mathematical combinations restricts the search space and often yields sub-optimal compromises. 
% Recent large language models (LLM)-based methods provide a different interface for scenario generation \cite{zhang2024chatscene,mei2025llm,tian2024enhancing}. However, these methods deploy LLM as a single-step generator or a substitute for a specific pipeline component, preventing them from iteratively searching for optimal solutions along the attack-realism frontier.

% \vspace{-3pt}
Existing safety-critical scenario generation methods attempt to manage this trade-off by assigning scalar weights \cite{nie2025steerable}, applying rule-based validity filters \cite{chen2024frea}, or restricting the search to learned traffic priors \cite{xu2025diffscene, liu2026adv}. While effective within specific pipelines, these methods rely heavily on designer-specified heuristics and handcrafted rules, confining the generation process to known priors and human intuition.
Consequently, they struggle to discover novel adversarial mechanisms and often collapse the inherently diverse attack-realism Pareto structure into a sub-optimal compromise. 
While recent large language model (LLM)-based methods provide a more flexible generation interface \cite{zhang2024chatscene,mei2025llm,tian2024enhancing}, they often deploy the LLM as a single-step generator or a static pipeline component, lacking the iterative evolutionary loop required to discover optimal solutions along the Pareto frontier.

% \vspace{-3pt}
Concurrently, recent open-ended agentic evolution systems have demonstrated strong capabilities in replacing fixed search rules with adaptive, LLM-driven outer loops \cite{shinn2023reflexion, wang2023voyager, lu2024ai, novikov2025alphaevolve}. Frameworks such as \cite{lange2025shinkaevolve,qu2026coral,liu2026evox} have achieved notable success in software development and algorithm discovery. However, while open-ended search can expand the generator design space, directly deploying them in closed-loop driving simulators is infeasible. 
They are usually evaluated through executable tests, automated reviewers, or scalar benchmark scores \cite{liu2026evox}. 
Without strict simulator grounding and domain-specific constraints, general agents tend to collapse the multi-objective tension into single-scalar maximization. Driven by a single target, an unconstrained agent can ``exploit'' the environment by generating physically impossible or implausible scenarios (e.g., overlapping vehicles) just to hack the adversarial score. Furthermore, the effectiveness of agentic evolution relies on extensive and unbounded trial-and-error, which is computationally prohibitive in physical simulators \cite{li2022metadrive,dosovitskiy2017carla}.

To bridge this gap, we present \method{}, the first fully automated, LLM-based multi-objective agentic evolution framework tailored for safety-critical autonomous driving. Instead of collapsing the objectives into a predefined scalar score, \method{} maintains a Pareto archive to preserve diverse trade-offs between attack and realism. To prevent unconstrained exploitation, we introduce a simulator-grounded, role-specialized actor-critic agent architecture to navigate the search space. Specifically, a memory-driven actor utilizes rich structural context to propose bounded improvements to the generator, while a cascade of critic agents evaluates structural validity and physical realism to filter out implausible proposals before simulation. Finally, to handle the high computational cost of closed-loop rollouts, a world evaluator routes promising candidates. By continually maintaining an archive of labeled candidates, \method{} preserves the best trade-offs and guides future evolutionary generations through structured feedback.
Our main contributions are threefold:
\begin{itemize}[leftmargin=*, itemsep=0pt, topsep=0pt]
    \item We introduce \method{}, the first agentic evolution framework for safety-critical driving scenario generation, enabling multi-objective iterative optimization and bypassing handcrafted heuristics.
    \item We design a simulator-grounded actor-critic proposal mechanism that drives autonomous exploration under explicit constraints, complemented by a self-evolving world evaluator that can learn from simulation feedback to prioritize promising proposals and reduce computational costs.
    \item Extensive experiments on MetaDrive and CARLA demonstrate that \method{} not only significantly expands the attack-realism Pareto frontier across various baseline generators, but also produces highly valuable scenarios that improve downstream policy training performance.
\end{itemize}

%===============================================================================

\vspace{-7pt}
\section{Preliminaries and Problem Formulation}
\label{sec:preliminaries}

\vspace{-8pt}
\paragraph{Problem setting.}
We study adversarial scenario generation in simulations. Let $\egopolicy$ be a fixed ego driving policy, $\seedset$ a pool of scenarios, and $\simulator$ a closed-loop simulator. A scenario $s \in \seedset$ specifies the map context, non-ego agents, initial states, and environment metadata. A generator candidate $c$ does not modify $\egopolicy$ but proposes simulator-executable interventions that alter non-ego behavior or scene conditions, such as interaction timing, motion perturbation, or behavioral patterns. 
Formally, $c$ induces a distribution over compilable interventions through scene-conditioned latent decisions $\omega$:
% a generator $c$ induces scene-conditioned latent decisions $\omega$ and compiles them into simulator operators:
\begin{equation}
    p_c(s,\omega,\zeta,\tau,y)
    =
    \mu_{\seedset}(s)\,
    q_c\bigl(\omega\mid \psi(s)\bigr)\,
    \mathbf{1}\{\zeta=\compiler(c,\omega;s)\}\,
    p_{\simulator}\bigl(\tau\mid s,\zeta,\egopolicy\bigr)\,
    p(y\mid\tau),
    \label{eq:prelim-generator-factorization}
\end{equation}
where $\mu_{\seedset}$ is the fixed scenario seed distribution, $\psi(s)$ denotes scene features observed by the generator, $\omega$ contains scene-level choices such as target actor, interaction window, motion parameters, and realism guards, $\zeta\in\operators$ is the compiled intervention, $\tau(c,s)$ denotes a rollout sampled from $p_{\simulator}(\cdot\mid s,\zeta,\egopolicy)$ after applying $\zeta=\compiler(c,\omega;s)$, and $y$ denotes rollout outcomes used to compute statistics. 
The ego policy $\egopolicy$, seed pool, and simulator dynamics are fixed. \method{} \textit{evolves the reusable generator program $c$, namely the conditional proposal $q_c(\omega\mid\psi(s))$ and the compiler-mediated intervention map $\compiler(c,\omega;s)$ that projects the candidate to simulator operators}. 
Thus, at the \textbf{generator level}, $c$ specifies a reusable adversarial mechanism across scenes, while at the \textbf{scene level}, $(\omega,\zeta)$ realizes one admissible simulator intervention for a particular scene or a scene family.
% In both cases, a candidate must be represented as structured simulator operators before it can be evaluated.

\vspace{-8pt}
\paragraph{Objective and search space.}
A useful adversarial scenario should expose safety-relevant failures while remaining physically and semantically plausible. For a rollout $\tau(c,s)$, let $a(\tau) \in [0,1]$ measure adversariality, such as collision exposure or near-miss severity, and let $r(\tau) \in [0,1]$ measure realism, such as map compliance and kinematic feasibility. Over an evaluation subset $\seedset_{\mathrm{eval}}$, we use
\begin{equation}
    A(c)=\frac{1}{|\seedset_{\mathrm{eval}}|}\sum_{s\in\seedset_{\mathrm{eval}}} a\bigl(\tau(c,s)\bigr),
    \qquad
    R(c)=\frac{1}{|\seedset_{\mathrm{eval}}|}\sum_{s\in\seedset_{\mathrm{eval}}} r\bigl(\tau(c,s)\bigr),
    \label{eq:prelim-objectives}
\end{equation}
with larger values preferred. The central difficulty is that these two objectives are coupled but not identical: maximizing attack alone can create implausible scenes, and optimizing realism alone may leave the ego policy unstressed.
We thus treat the generation as a \textbf{multi-objective search problem} over $F(c)=(A(c),R(c))$. 
Candidate search is restricted to the simulator-admissible set
\begin{equation}
    \mathcal{C}_{\mathrm{sim}}=
    \left\{c\in\mathcal{C}\;\middle|\;\validator(c)=1,\;\compiler(c)\in\operators,\;\Gamma(c,s)=1\;\text{for evaluated seeds }s\right\},
    \label{eq:prelim-admissible}
\end{equation}
where $\validator$ is a deterministic validator and $\Gamma$ checks scene-specific map, dynamics, and simulator constraints. A candidate $c_i$ Pareto-dominates $c_j$, written $c_i \succ c_j$, if $A(c_i)\ge A(c_j)$, $R(c_i)\ge R(c_j)$, and at least one inequality is strict. These definitions motivate \method: evolve generator mechanisms through bounded local edits, evaluate accepted candidates with real simulator rollouts, and maintain an archive of non-dominated trade-offs, to expand this attack-realism frontier.

\vspace{-10pt}
\section{EvoDrive}
\label{sec:method}

\vspace{-15pt}
\begin{figure}[!htbp]
\centering
\includegraphics[width=1\linewidth]{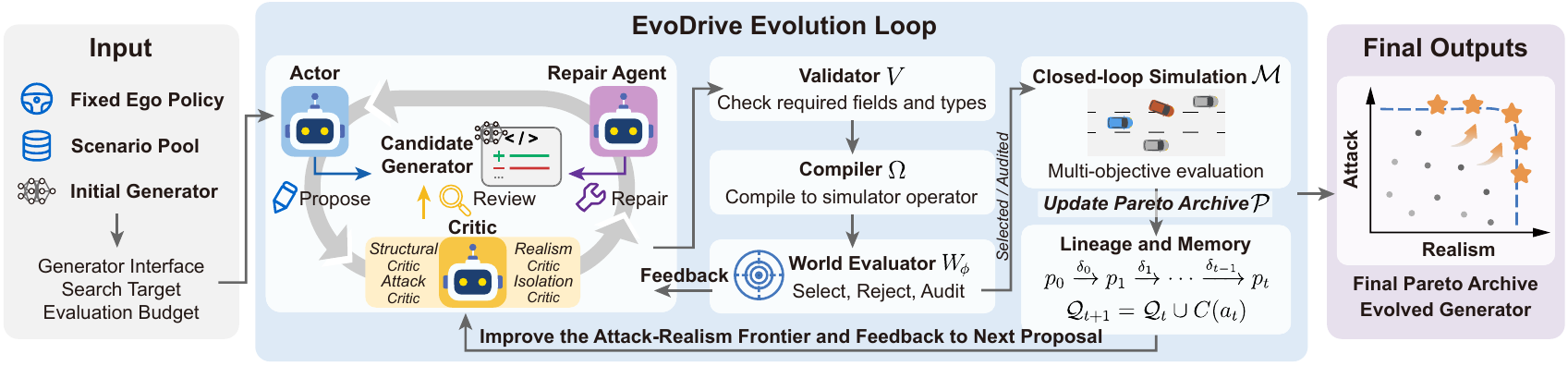}
\vspace{-15pt}
\caption{Conceptual architecture of \method{}. 
Actor-critic LLM agents iteratively propose, verify, simulate, and archive generators to improve the attack-realism frontier.}
\label{fig:method-architecture}
\end{figure}

\vspace{-15pt}
\subsection{Overview}
\label{subsec:method-overview}
\vspace{-5pt}
\method{} is an LLM-based agentic evolution framework for discovering adversarial but realistic traffic scenarios. {Given a fixed ego policy $\egopolicy$, a scenario pool $\seedset$, initial generator programs, and an evaluation budget, \method{} searches for generators that improve the attack-realism frontier. The interface registry specifies the editable fields, admissible values, validation checks, and compilation rules for each generator family.}
{The output is a Pareto archive of candidates, together with their simulator evidence, reusable memories, and lineage records describing parent-child edits and evaluation outcomes.}
Following Eq.~\eqref{eq:prelim-generator-factorization}, an evolution changes $q_c(\omega\mid\psi(s))$ and $\compiler(c,\omega;s)$, while the seed distribution, ego policy, simulator transition, and metrics remain locked. 
Thus, \method{} evolves adversarial generation itself without modifying the driving policy or perturbing a single logged trajectory.
An evolution epoch follows the sequence:
\begin{equation}
    \text{parent generator}
    \rightarrow \text{local edit}
    \rightarrow \validator
    \rightarrow \compiler
    \rightarrow \world
    \rightarrow \simulator
    \rightarrow \archive
    \rightarrow \text{memory and lineage}.
    \label{eq:method-pipeline}
\end{equation}
LLM-based coding agents~\cite{openai_codex_2026,anthropic_claude_code_2026} first suggest bounded edits to a selected parent generator, using retrieved memory and the current search target as context. Critic agents and deterministic validators then check whether the edited generator is well-formed, belongs to the allowed interface, and can be compiled into simulator operators. A calibrated world evaluator $\world$ routes accepted candidates by prioritizing promising or uncertain ones before expensive rollout.
\method{} separates three roles that are often conflated in agentic optimization. Proposal agents generate hypotheses about how to modify a generator; deterministic components enforce admissibility and simulator compatibility; and closed-loop simulation with Pareto archiving decides whether the resulting candidate improves the frontier. Figures~\ref{fig:method-architecture} and~\ref{fig:system-arch} illustrate the architecture and organization that realize this design.

\vspace{-8pt}
\subsection{Evolvable Generator Representation}
\label{subsec:method-generator-programs}

\vspace{-8pt}
\textbf{Generator state} is the object that is edited, compiled, and evaluated. Instead of free-form code, \method{} represents each evolvable generator as a typed state structure at each iteration $t$
\begin{equation}
    p_t = (b, \theta_t, G_t, \rho_t, H_t, \kappa_t, \alpha_t),
    \label{eq:method-program-state}
\end{equation}
where $b$ is the parent baseline, $\theta_t$ are bounded parameters, {$G_t$ is an operator graph combining supported simulator operations such as actor selection, interaction timing, and motion changes}, $\rho_t$ is a scenario-conditioned policy for choosing operators, {$H_t$ is a set of reusable mechanism modules that express rules for traffic interactions through these operators}, $\kappa_t$ specifies the required constraints, and $\alpha_t$ records the ancestor chain used to audit parent-child continuity.
% The conditioned policy allows the same generator to use different settings for different scenes while remaining reusable.
$\rho_t$ selects settings from $G_t$ according to scene features, allowing the same generator to use different interventions in different scenes while preserving a shared mechanism.
Equivalently, $p_t$ parameterizes both the conditioned proposal $q_{p_t}(\omega\mid\psi(s))$ and the compilation rule $\compiler(p_t,\omega;s)$ in Eq.~\eqref{eq:prelim-generator-factorization}. This makes the editable state broader than a scalar hyperparameter vector, but more auditable than unrestricted code generation.

\vspace{-8pt}
\paragraph{Generator interfaces.}
$p_t$ stores the current values of a generator, whereas an interface specifies the action domains and checks for such states.
To keep the proposals from agents feasible in the search space, each generator family exposes an interface $\mathcal{I}_b$ that defines the \textit{admissible state space, edit field, and compilation contract} of evolution. 
It prevents an agent from editing the ego policy, changing the benchmark, bypassing the evaluator, or introducing arbitrary code. 
Formally, for source family $b$, we write
    $\mathcal{I}_b =
    \left(
    \Theta_b,\,
    \mathcal{G}_b,\,
    {\mathcal{R}_b,\,}
    \mathcal{H}_b,\,
    {\mathcal{F}_b,\,}
    V_b,\,
    \compiler_b
    \right)$,
where $\Theta_b$ is the admissible parameter domain, $\mathcal{G}_b$ is the operator-graph family, {$\mathcal{R}_b$ is the admissible class of scenario-conditioned policies,} $\mathcal{H}_b$ is the mechanism vocabulary, {$\mathcal{F}_b$ is the family-specific constraint schema,} $V_b$ is the validator, and $\compiler_b$ is the family-specific compiler.
$p_t$ is admissible for family $b$ only when $\theta_t\in\Theta_b$, $G_t\in\mathcal{G}_b$, $\rho_t\in\mathcal{R}_b$, $H_t\subseteq\mathcal{H}_b$, $\kappa_t\in\mathcal{F}_b$, and $V_b(p_t)=1$.
$\mathcal{I}_b$ specifies not only what to edit, but also how it is checked and converted into executable interventions, bounding where evolution may move.

% \paragraph{Bounded generator edits.}
% A proposal actor edits a parent generator locally.
% We denote the edit by $\delta_t$ and restrict its support to the generator fields that affect scenario generation:
% \begin{equation}
%     \mathrm{supp}(\delta_t) \subseteq
%     \left\{
%     \theta_t,\;
%     G_t,\;
%     \rho_t,\;
%     H_t
%     \right\}.
%     \label{eq:method-edit-surface}
% \end{equation}
% The child is produced by a deterministic transition $p_{t+1}=T_b(p_t,\delta_t)$ only if the edit belongs to the same source family and stays inside $\mathcal{I}_b$. 
% % \methodadd{The transition also checks that the edit was generated for the exact parent hash $h_t$, so a stale or mismatched patch cannot be applied to another lineage state.}
% $T_b$ also verifies parent identity, so a lineage-incompatible edit cannot be applied to a different generator state.
% This local edit makes the search traceable and prevents the agent from replacing a baseline with unrelated code or modifying protected evaluation.

% \vspace{-3pt}
\textbf{Bounded generator edits} are the transition objects that move one valid state to another one under $\mathcal{I}_b$.
A proposal agent edits a {valid} parent generator locally.
We denote the edit by $\delta_t$ and restrict its support to the generator fields that affect scenario generation:
$\mathrm{supp}(\delta_t) \subseteq
    \left\{
    \theta_t,\;
    G_t,\;
    \rho_t,\;
    H_t
    \right\}$.
% \methodadd{The protected fields $b$, $\kappa_t$, and $\alpha_t$ are outside the actor-proposed support: the source family and required constraints are inherited from the parent and interface, while the lineage record is updated deterministically.}
The child is produced by a deterministic transition $p_{t+1}=T_b(p_t,\delta_t)$ only if the edit belongs to the same source family and stays inside $\mathcal{I}_b$.
% \methodadd{The transition also checks that the edit was generated for the exact parent hash $h_t$, so a stale or mismatched patch cannot be applied to another lineage state.}
$T_b$ also verifies parent identity, so a lineage-incompatible edit cannot be applied to a different generator state.
{Operationally, $T_b$ applies the local edit, recomputes lineage fields, and accepts the child only if $V_b(p_{t+1})=1$.}
{Therefore, $\mathcal{I}_b$ defines the legal region, $\delta_t$ proposes one move inside that region, and $T_b$ instantiates an auditable parent-child transition.}

\vspace{-8pt}
\paragraph{Compilation.}
After validation, the compiler converts a scene-level decision $\omega$ into an intervention
$\zeta=\compiler_b(p_{t+1},\omega;s)=
    (
    O_{\mathrm{select}},
    O_{\mathrm{actor}},
    O_{\mathrm{timing}},
    O_{\mathrm{motion}},
    O_{\mathrm{search}},
    O_{\mathrm{filter}},
    O_{\mathrm{div}};
\Gamma_{\mathrm{map}},\Gamma_{\mathrm{dyn}},\Gamma_{\mathrm{api}}
    ),$ where the components select applicable scenes, choose non-ego actors, set interaction windows, instantiate motion changes, search bounded parameters, remove unrealistic rollouts, and maintain scenario diversity. $\Gamma_{\mathrm{map}}$, $\Gamma_{\mathrm{dyn}}$, and $\Gamma_{\mathrm{api}}$ denote map, dynamics, and interface constraints. 
{Compilation is the bridge between an abstract adversarial mechanism and closed-loop physical simulation: a proposal is admissible only if it can be expressed through these constrained simulator operators.}

\vspace{-5pt}
\subsection{Actor-Critic Agents for Specialized Proposal and Repair}
\label{subsec:method-agents-validation}

\vspace{-8pt}
\paragraph{Agent roles as proposal operators.}
Let $x_t$ denote the context for iteration $t$, including the parent summary, the generator interface, recent memory, feedback from prior attempts, and the current search target.
The target $\eta_t$ may be to increase attack at high realism, diversify the frontier, or improve a trade-off.
\method{} introduces an actor-critic decomposition at the proposal stage.
To avoid overloading the scenario-conditioned policy $\rho_t$, we denote role-specialized LLM agents by
\begin{equation}
{
    \mathcal{A}
    =
    \big\{
    \mathsf{A}_{\mathrm{actor}},
    \mathsf{A}_{\mathrm{struct}},
    \mathsf{A}_{\mathrm{attack}},
    \mathsf{A}_{\mathrm{real}},
    \mathsf{A}_{\mathrm{iso}},
    \mathsf{A}_{\mathrm{repair}}
    \big\}.
}
\end{equation}
For a selected parent $p_t$, the program actor produces an initial edit $\delta_t^{(0)}
    \sim
    \mathsf{A}_{\mathrm{actor}}
    \bigl(\cdot \mid p_t,x_t,\eta_t\bigr)$.
The actor also returns a short mechanistic rationale explaining why the edit is expected to change attack-realism behavior.
When multiple actor specializations are used, they may emphasize attack pressure, realism preservation, or portfolio selection, but their outputs share the same interface.

% \paragraph{Agent roles as proposal operators.}
% Let $x_t$ denote the context for iteration $t$, including the parent summary, the generator interface, recent memory, feedback from prior attempts, and the current search target. 
% The current target $\eta_t$ may be to increase attack at high realism, diversify the frontier, or improve a parent trade-off. 
% \method{} instantiates a set of role-specialized proposal agents:
% \begin{equation}
%     \mathcal{R}=\big\{\rho_{\mathrm{actor}},\rho_{\mathrm{struct}},\rho_{\mathrm{attack}},
%     \rho_{\mathrm{real}},\rho_{\mathrm{iso}},\rho_{\mathrm{repair}}\big\}.
% \end{equation}
% For a selected parent $p_t$, the program actor $\rho_{\mathrm{actor}}$ is instructed to propose a candidate edit $\delta_t^{(0)}\sim \rho_{\mathrm{actor}}(\delta \mid p_t,x_t,\eta_t)$. The actor agent also returns a short causal hypothesis explaining why the edit should change the attack--realism behavior.
% In practice, different actor roles may emphasize attack pressure, realism preservation, or portfolio selection, but their outputs share the same interface.

\vspace{-8pt}
\paragraph{Critic-gated repair.}
The proposal is reviewed before simulations by critics with complementary inductive biases.
A structural critic {$\mathsf{A}_{\mathrm{struct}}$} checks whether the edit is well-formed and uses supported fields.
An attack critic {$\mathsf{A}_{\mathrm{attack}}$} evaluates whether the edit plausibly increases adversarial pressure.
A realism critic $\mathsf{A}_{\mathrm{real}}$ checks physical continuity, map topology, and traffic plausibility.
An isolation critic $\mathsf{A}_{\mathrm{iso}}$ verifies that the proposal remains within the generator family and does not use protected information.
The critic cascade returns an advisory decision in $\{\mathrm{approve},\mathrm{repair},\mathrm{reject}\}$.
If $\mathrm{repair}$ is requested, the repair agent receives the critic feedback and produces a revised edit $\delta_t^{(m+1)}
    \sim
    \mathsf{A}_{\mathrm{repair}}
    \bigl(\cdot \mid x_t,\delta_t^{(m)},r_t^{(m)}\bigr)$,
where $r_t^{(m)}$ is the collection of feedback.
The loop terminates when the edit passes the checks, is rejected, or reaches a budget.
This loop is denoted as:
\begin{equation}
\begin{aligned}
    (p_t,x_t,\eta_t)
    &\xrightarrow{\mathsf{A}_{\mathrm{actor}}}
    \delta_t^{(0)}
    \xrightarrow[\text{repair by } \mathsf{A}_{\mathrm{repair}}]
    {\mathsf{A}_{\mathrm{struct}},\mathsf{A}_{\mathrm{attack}},\mathsf{A}_{\mathrm{real}},\mathsf{A}_{\mathrm{iso}}}
    \delta_t
    \xrightarrow{T_b}
    c_t .
\end{aligned}
    \label{eq:method-role-loop}
\end{equation}
The resulting $c_t$ is accepted for evaluation after validation.
This decomposition separates evolutionary search from verification. The actor is encouraged to exploit weaknesses of the ego policy, whereas the critics induce constraints that are often in tension with pure attack maximization. Therefore, the resulting proposal is a structured hypothesis about how to move the attack-realism frontier, which differentiates \method{} from general agentic search frameworks.

\vspace{-8pt}
\paragraph{Deterministic validation.}
After the role loop, \method{} applies a deterministic validator:
\begin{equation}
    \validator(c_t)
    =
    \prod_{j\in\mathcal{J}} V_j(c_t),
    \qquad
    \mathcal{J}
    =
    \{
    \mathrm{schema},
    \mathrm{parent},
    \mathrm{interface},
    \mathrm{isolation},
    \mathrm{safety},
    \mathrm{compile}
    \},
    \label{eq:method-validator}
\end{equation}
where each $V_j(c_t)\in\{0,1\}$.
The validator checks required fields and types, such as candidate schema, parent identity, allowed edits, protected resources, and compile feasibility. 
If any factor is zero, the candidate is recorded as invalid, and no simulator rollout is launched. 
Consequently, LLM agents function as hypothesis generators inside a controlled environment. 
% and are forbidden to change the official evaluator, rewrite the benchmark split, or substitute metrics.

\vspace{-5pt}
\subsection{World-Conditioned Candidate Routing and Evaluator Evolution}
\label{subsec:world}
\vspace{-5pt}
\paragraph{World evaluator.}
Closed-loop simulator rollouts are expensive, particularly when many proposals are generated per epoch.
\method{} introduces a calibrated \textit{world evaluator} $W_\phi$ as a pre-rollout routing model. It prioritizes which candidate-scenario pairs should receive simulator budget before labels are obtained from $\mathcal{M}$. 
For a pair $(c,s)$, it predicts
$
    W_\phi(c,s)
    =
    \big(\widehat{A}_\phi(c,s),\widehat{R}_\phi(c,s),\widehat{U}_\phi(c,s)\big),
$ where $\widehat{A}_\phi$ estimates attack potential, $\widehat{R}_\phi$ estimates realism, and $\widehat{U}_\phi$ estimates uncertainty.
The input features are computed from the candidate state, scene features, ego-policy metadata, interaction summaries, counterfactual probes, and calibration statistics from historical simulations.
Across a routing subset {$\mathcal{S}^{\mathrm{route}}_t\subseteq\seedset$}, the world evaluator aggregates scene-level predictions into
$
    \widehat{\mathbf{F}}_\phi(c)
    =
    \frac{1}{|\mathcal{S}^{\mathrm{route}}_t|}
    \sum_{s\in\mathcal{S}^{\mathrm{route}}_t}
    \big(
    \widehat{A}_\phi(c,s),
    \widehat{R}_\phi(c,s)
    \big),
    \widehat{U}_\phi(c)
    =
    \operatorname{Agg}_{s\in\mathcal{S}^{\mathrm{route}}_t}
    \widehat{U}_\phi(c,s),$ giving a candidate-level estimate of expected attack-realism behavior and uncertainty before expensive closed-loop evaluation.

% \subsection{World-Conditioned Candidate Routing and Evaluator Evolution}
% \label{subsec:world}

% \paragraph{World evaluator.}
% Closed-loop simulator rollouts are expensive, particularly when many proposals are generated per epoch. \textsc{EvoDrive} introduces a simulator-calibrated \textit{world evaluator}  $W_\phi$ as a routing model before labels are obtained from $\mathcal{M}$. 
% For a candidate-scenario pair $(c,s)$, it predicts
% \begin{equation}
%     W_\phi(c,s) = \big(\widehat{A}_\phi(c,s),\widehat{R}_\phi(c,s),\widehat{U}_\phi(c,s)\big),
% \end{equation}
% where $\widehat{A}$ estimates attack potential, $\widehat{R}$ estimates realism, and $\widehat{U}$ estimates uncertainty.
% The input features are computed from the candidate program, ego-policy metadata, synthetic interactions, counterfactual branches, and calibration statistics from historical simulations. 
% Across a subset $\mathcal{S}_t$, the world evaluator aggregates predictions into $\widehat{F}_\phi(c)=\mathbb{E}_{s\in\mathcal{S}_t}[\widehat{A}_\phi(c,s),\widehat{R}_\phi(c,s)]$ \methodadd{and computes candidate-level uncertainty $\widehat{U}_\phi(c)$ from the routed seed predictions.}

\vspace{-5pt}
\paragraph{Routing decisions.}
Given thresholds $\lambda_A,\lambda_R,\lambda_U$, the world evaluator produces a routing decision:
\begin{equation}
{
    d_\phi(c)
    =
    D\bigl(
    \widehat{\mathbf{F}}_\phi(c),
    \widehat{U}_\phi(c);
    \lambda_A,\lambda_R,\lambda_U
    \bigr)
    \in
    \{\mathrm{select},\mathrm{reject},\mathrm{audit}\}.
}
    \label{eq:method-routing}
\end{equation}
Candidates predicted to be promising and certain enough are selected for real rollouts. Those predicted to be implausible or uninformative may be rejected before simulation.
The audit decision forces real evaluation for high-uncertainty or monitoring cases despite a weak predicted score.
Audit labels are essential because they expose false rejections and keep the evaluator calibrated under distribution shift.
This routing policy changes the priority and allocation of rollout budgets.
After evaluation, simulator labels are added to a calibration set
$
    \mathcal{D}^{W}_t
    =
    \left\{\bigl({\xi_i}, y_i\bigr)\right\}_{i=1}^{n_t},
    y_i
    =
    \bigl(A_i,R_i,Q_i\bigr),
$ where $\xi_i$ contains the candidate and routing features, and $y_i$ is computed from simulator rollouts.
The auxiliary term $Q_i$ stores rollout diagnostics to calibrate routing and monitoring.
These statistics are then used by later epochs and by the nested evaluator-evolution loop described below.

% \paragraph{Routing decisions.}
% Given thresholds $\lambda_A,\lambda_R,{\lambda_U}$, the world evaluator produces a routing decision:
% \begin{equation}
%     d_\phi(c) \in \{\mathrm{select},\mathrm{reject},\mathrm{audit}\},
%     \qquad
%     d_\phi(c)=D\bigl({\widehat{A}_\phi(c)},{\widehat{R}_\phi(c)},{\widehat{U}_\phi(c)};\lambda_A,\lambda_R,{\lambda_U}\bigr).
% \end{equation}
% Candidates predicted to be promising and sufficiently certain are selected for real rollouts. Candidates predicted to be implausible or uninformative may be rejected before simulation.
% The audit decision forces real evaluation for high-uncertainty or monitoring cases despite a weak predicted score. Audit labels are essential because they expose false rejections and keep the evaluator calibrated under distribution shift.
% This routing policy changes the priority and allocation of rollout budgets.
% After selected candidates are evaluated, real labels are added to a calibration set
% \begin{equation}
%     \mathcal{D}^{W}_t = \left\{\bigl(z_i, y_i\bigr)\right\}_{i=1}^{n_t},
%     \qquad
%     y_i = \bigl(A_i,R_i,Q_i\bigr),
% \end{equation}
% where $z_i$ contains the candidate and features, and $y_i$ is computed from simulator rollouts. 
% \methodadd{The auxiliary term $Q_i$ stores rollout diagnostics such as invalid-scene status, route completion, or diversity metadata when these quantities are available.}
% % Calibration estimates prediction error, ranking consistency, and uncertainty failures. 
% These statistics are then used by later epochs and by the nested evaluator-evolution loop described below.

\vspace{-5pt}
\paragraph{Evaluator self-evolution.}
A static routing model can be inaccurate as the generator distribution shifts.
\method{} thus treats the evaluator snapshot as an evolvable object.
Let $\mathcal{Z}^{W}_t$ summarize prediction errors, false rejects, uncertainty failures, and ranking disagreements.
Once sufficient labels are available, a world agent proposes a bounded evaluator update $\Delta\phi_t
    \sim
    \mathsf{A}_{W}
    \bigl(\cdot \mid \phi_t,\mathcal{D}_t^W,\mathcal{Z}^{W}_t,\mathcal{U}_W\bigr)$,
where $\mathcal{U}_{W}$ includes the allowed updates for calibration, ranking, uncertainty, and prescreen.
The proposed snapshot $\phi'_t=\mathrm{Apply}_W(\phi_t,\Delta\phi_t)$ is promoted only if a deterministic judge $J_W(\phi_t,\phi'_t,\mathcal{D}_t^W)$ accepts it.
% \begin{equation}
%     \phi_{t+1}=
%     \begin{cases}
%     \phi'_t, & J_W(\phi_t,\phi'_t,\mathcal{D}_t^W)=1,\\
%     \phi_t, & \text{otherwise.}
%     \end{cases}
% \end{equation}
% The judge evaluates the proposed snapshot only on simulator-labeled evidence and checks that routing quality improves without introducing monitored calibration failures.
Promoted snapshots are then used by later epochs for routing and audit.
They do not overwrite simulator labels or directly promote candidates to the Pareto archive.

% Promoted snapshots are then used by later generator epochs for routing and audit. \methodadd{They do not overwrite simulator labels and cannot directly promote generator candidates to the archive.}

% \paragraph{Evaluator self-evolution.}
% A static surrogate can become inaccurate as the generator distribution shifts. \method{} therefore treats the evaluator snapshot as an evolvable object.
% Let {$\mathcal{Z}^{W}_t$} summarize prediction errors, false rejects, uncertainty failures, and ranking disagreements.
% Once sufficient labels are available, a \textit{world agent} is instructed to propose a bounded evaluator update
% \begin{equation}
%     \Delta\phi_t \sim \rho_W\big(\phi_t,\mathcal{D}_t^W,{\mathcal{Z}^{W}_t},\mathcal{U}_W\big),
% \end{equation}
% where $\mathcal{U}_{W}$ includes configuration-level updates to calibration, ranking, and prescreen policies. The proposed snapshot $\phi'_t=\mathrm{Apply}_W(\phi_t,\Delta\phi_t)$ is promoted only if a deterministic judge $J_W$ accepts it:
% \begin{equation}
%     \phi_{t+1}=
%     \begin{cases}
%     \phi'_t, & J_W(\phi_t,\phi'_t,\mathcal{D}_t^W)=1,\\
%     \phi_t, & \text{otherwise.}
%     \end{cases}
% \end{equation}
% Promoted snapshots are then used by later generator epochs for routing and audit. \methodadd{They do not overwrite simulator labels and cannot directly promote generator candidates to the archive.}

\vspace{-5pt}
\subsection{Simulator Evaluation and Pareto Archive}
\label{subsec:method-pareto}

% \minew{
% \paragraph{Multi-objective evaluation.}
% A selected {\color{magenta}or audited} candidate $c$ is evaluated on $\mathcal{S}_{\mathrm{eval}}\subseteq\seedset$ through closed-loop simulator rollouts.
% For each candidate $c$, \method{} computes the attack and realism objectives in Eq.~\eqref{eq:prelim-objectives} and maintains an archive storing the non-dominated attack--realism frontier.
% Let $\history_t$ be the set of simulator-evaluated candidates available after epoch $t$.
% {\color{magenta}
% We use the standard maximization dominance relation: $c'\succ c$ if $A(c')\ge A(c)$ and $R(c')\ge R(c)$, with at least one strict inequality.
% }
% The frontier is
% }

\vspace{-5pt}
\paragraph{Multi-objective evaluation.}
A selected or audited candidate $c$ is evaluated on $\mathcal{S}_{\mathrm{eval}}\subseteq\mathcal{S}$ through simulator rollouts.
For each $c$, \method{} computes the objectives in Eq.~\eqref{eq:prelim-objectives} and maintains an archive storing the non-dominated attack-realism frontier. 
{In MetaDrive, realism aggregates map and route compliance, collision validity, and motion continuity. In CARLA, the adversarial realism score (ARS) combines kinematic, map, interaction, and distance consistency. Detailed definitions are provided in Appendices~\ref{app:objective-computation} and~\ref{app:carla-setup}.}
Let $\history_t$ be the set of {simulator-evaluated} candidates available after epoch $t$. The frontier is $\archive_t =
    \left\{c\in\history_t\;\middle|\;\nexists c'\in\history_t \text{ such that } c'\succ c\right\}$, which avoids reducing scenario generation to a single scalar reward and preserves different kinds of useful scenarios: high-attack candidates that remain plausible, high-realism candidates that still reveal weaknesses, and intermediate ones that improve coverage of the trade-off surface.

\vspace{-5pt}
\paragraph{Acceptance rules.}
Several gates are used to decide whether an evaluated candidate should be retained as an active successor for future search. The Pareto archive remains the non-dominated subset.
Frontier densification accepts $c$ when $c$ is non-dominated and increases hypervolume:
$\Delta\mathrm{HV}(c)
    =
    \mathrm{HV}\big({\archive_t}\cup\{c\}\big)
    -
    \mathrm{HV}({\archive_t})
    >
    0$,
with respect to a fixed reference point.
A local frontier push accepts a candidate that improves attack in a high-realism region or improves realism near a high-attack anchor:
$\big[R(c)\ge r_0 \land A(c)>\max_{c'\in{\archive_t}: R(c')\ge r_0} A(c')\big]
    \;\lor\;
    \big[A(c)\ge a_0 \land R(c)>\max_{c'\in{\archive_t}: A(c')\ge a_0} R(c')\big].$
The maxima are interpreted over the corresponding archive region; if the region is empty, the gate falls back to the non-dominance and hypervolume checks.
A parent trade-off push accepts a child relative to its parent $p$ when it improves one objective while bounding the degradation of the other:
$\big[A(c)>A(p) \land R(c)\ge R(p)-\epsilon_R\big]
    \;\lor\;
    \big[R(c)>R(p) \land A(c)\ge A(p)-\epsilon_A\big].$
These gates retain diverse trade-offs and avoid collapsing the search to a fixed scalarization.
They also allow locally useful descendants to remain available for future evolution. 
% even when the archive update is ultimately governed by Pareto dominance.

% \paragraph{Acceptance rules.}
% \method{} promotes a candidate when it advances the archive under one of several gates. Frontier densification accepts $c$ when $c$ is non-dominated and increases hypervolume:
% \begin{equation}
%     \Delta\mathrm{HV}(c)=\mathrm{HV}\big({\archive_t}\cup\{c\}\big)-\mathrm{HV}({\archive_t})>0.
% \end{equation}
% A local frontier push accepts a candidate that improves attack in a high-realism region or improves realism near a high-attack anchor:
% \begin{equation}
%     \big[R(c)\ge r_0 \land A(c)>\max_{c'\in{\archive_t}: R(c')\ge r_0} A(c')\big]
%     \;\lor\;
%     \big[A(c)\ge a_0 \land R(c)>\max_{c'\in{\archive_t}: A(c')\ge a_0} R(c')\big].
% \end{equation}
% A parent trade-off push accepts a child relative to its parent $p$ when it improves one objective while bounding degradation of the other:
% \begin{equation}
%     \big[A(c)>A(p) \land R(c)\ge R(p)-\epsilon_R\big]
%     \;\lor\;
%     \big[R(c)>R(p) \land A(c)\ge A(p)-\epsilon_A\big].
% \end{equation}
% These gates retain diverse trade-offs and avoid collapsing the search to a fixed scalarization.

\vspace{-5pt}
\paragraph{Lineage, memory, and parallelization.}
Each proposal attempt $a_t$ is attached to a lineage record. Accepted children update the active generator, while rejected children are retained as failed proposals and negative evidence, yielding an auditable chain $p_0 \xrightarrow{\delta_0} \cdots \xrightarrow{\delta_{t-1}} p_t$. After each attempt, a curator compresses simulator outcomes and feedback into typed memory cards, $\mathcal{Q}_{t+1}=\mathcal{Q}_t \cup C(a_t)$, including failure memories, success motifs, notes, and reflection plans. Memory retrieval is causal, $\mathcal{K}_{t+1}=\mathrm{Retrieve}(\mathcal{Q}_{\le t})$, and cross-lineage sharing is restricted to mechanism-level motifs rather than implementation details. To scale exploration, \method{} may generate $K$ candidates from the same immutable epoch snapshot. Workers share the same context but cannot mutate the archive, lineage store, or shared memory. A serial acceptor then applies validation, routing, selection, and updates after an epoch barrier, increasing proposal diversity while preserving a single deterministic transition order. Detailed schemas are given in Appendix~\ref{app:program-objects}, Appendix~\ref{app:archive-memory}, and Appendix~\ref{app:parallel-epochs}.

\vspace{-10pt}
\section{Experiment}
\label{sec:experiment}
\setlength{\textfloatsep}{8pt plus 1pt minus 1pt}
\setlength{\intextsep}{6pt plus 1pt minus 1pt}
\setlength{\floatsep}{6pt plus 1pt minus 1pt}
\setlength{\dbltextfloatsep}{8pt plus 1pt minus 1pt}
\setlength{\dblfloatsep}{6pt plus 1pt minus 1pt}

\vspace{-5pt}
To evaluate \method{}, we organize experiments around three questions: (1) \textit{whether evolved generators improve the attack-realism frontier}, (2) \textit{whether self-evolving agents discover informative mechanisms beyond scalar reparameterization}, and (3) \textit{whether the resulting scenarios remain useful for improving downstream policies}.
This section presents aggregated comparisons and evidence supporting main findings.
Appendix~\ref{app:experimental-setups} and~\ref{app:supplementary-results} provide the evaluation details and additional results.

\vspace{-10pt}
\paragraph{{Evaluation protocol.}}
{MetaDrive uses 500 scenes: 240 for generator evolution, 160 additional scenes that expand downstream SAC training to 400 scenes, and 100 for held-out evaluation. PF-Area@3 reports the percentage increase in attack-realism hypervolume over the parent frontier after adding up to three evolved descendants. After evolution, the active generator and these descendants are selected using evolution pool results and frozen before evaluation on the 100 held-out scenes.}
{LLMs, agent configuration, and computational cost are reported in Appendix~\ref{app:llm-config-cost}.}

\vspace{-3pt}
\subsection{Generator Evolution Across Driving Simulators}
\label{sec:exp-generator-evolution}

\begin{table*}[!htbp]
\vspace{-6pt}
\centering
\small
\setlength{\tabcolsep}{5pt}
\renewcommand{\arraystretch}{1.0}
\caption{Generator evolution in MetaDrive. Attack is the target collision rate; Realism penalty (RP) is $100(1-R)$; {PF-Area@3 is the hypervolume gain over the parent frontier with up to 3 descendants.}
% For RP change, negative values indicate a lower penalty.
}
\label{tab:metadrive-main-generator-evolution}
\resizebox{0.8\linewidth}{!}{
\begin{tabular}{lccccccc}
\toprule
 & \multicolumn{2}{c}{Parent} & \multicolumn{2}{c}{w/ \method} & \multicolumn{3}{c}{Change} \\
\cmidrule(lr){2-3}\cmidrule(lr){4-5}\cmidrule(l){6-8}
Method & Attack$\uparrow$ & RP$\downarrow$ & Attack$\uparrow$ & RP$\downarrow$ & Attack$\uparrow$ & RP$\downarrow$ & PF-Area@3$\uparrow$ \\
\midrule
CAT~\cite{zhang2023cat} & 0.345 & 10.66\% & 0.330 & \cellcolor{green!10}3.85\% & -4.3\% & \cellcolor{green!12}-6.81pp & \cellcolor{green!12}+7.4\% \\
ADV-BMT~\cite{liu2026adv} & 0.260 & 2.27\% & \cellcolor{green!10}0.295 & \cellcolor{green!10}1.77\% & \cellcolor{green!12}+13.5\% & \cellcolor{green!12}-0.51pp & \cellcolor{green!12}+18.7\% \\
AT~\cite{zhang2022adversarial} & 0.231 & 0.09\% & \cellcolor{green!10}0.265 & 1.06\% & \cellcolor{green!12}+14.7\% & +0.97pp & \cellcolor{green!12}+14.9\% \\
SAGE~\cite{nie2025steerable} & 0.211 & 0.06\% & \cellcolor{green!10}0.234 & 0.72\% & \cellcolor{green!12}+10.7\% & +0.66pp & \cellcolor{green!12}+13.2\% \\
LLM-Attacker~\cite{mei2025llm} & 0.221 & 0.28\% & \cellcolor{green!10}0.231 & 0.51\% & \cellcolor{green!12}+4.5\% & +0.23pp & \cellcolor{green!12}+5.2\% \\
KING~\cite{hanselmann2022king} & 0.261 & 0.16\% & \cellcolor{green!10}0.280 & 0.78\% & \cellcolor{green!12}+7.2\% & +0.62pp & \cellcolor{green!12}+7.0\% \\
SEAL~\cite{stoler2025seal} & 0.211 & 0.07\% & \cellcolor{green!10}0.245 & 1.02\% & \cellcolor{green!12}+16.0\% & +0.95pp & \cellcolor{green!12}+15.9\% \\
STRIVE~\cite{rempe2022generating} & 0.226 & 0.04\% & \cellcolor{green!10}0.245 & 0.74\% & \cellcolor{green!12}+8.3\% & +0.70pp & \cellcolor{green!12}+8.0\% \\
\bottomrule
\end{tabular}
}

\end{table*}
\paragraph{MetaDrive.}
We first evaluate \method{} by evolving representative scenario generators and measuring the resulting attack-realism trade-off in MetaDrive~\cite{li2022metadrive}.
Table~\ref{tab:metadrive-main-generator-evolution} reports the generator-level summary averaged over IDM and RL ego policies.
\method{} increases the Pareto area for every generator family, indicating that the evolved generators expand the attainable frontier. % rather than improving a single score in isolation.
The gains are complementary across generators, e.g., CAT primarily improves realism while preserving frontier coverage, whereas ADV-BMT improves active attack, realism penalty, and frontier area simultaneously.
% Full profiles are detailed in Appendix~\ref{app:supp-metadrive-policy}. Ablation studies in Appendix~\ref{app:supp-metadrive-diagnostics} further isolate the role of our evolution objective and architecture.
Full profiles and ablations on our objective and architecture are in App.~\ref{app:supp-metadrive-policy} and \ref{app:supp-metadrive-diagnostics}.

\vspace{-8pt}
\paragraph{{Realism validation.}}
{To evaluate realism against naturalistic traffic, we compare held-out rollouts with maneuver-matched naturalistic trajectories after evolution. Table~\ref{tab:realism-validation} reports motion and interaction Jensen-Shannon divergence (JSD) and out-of-support (OOS) rate. These metrics are used only for post-evolution evaluation. Averaged across eight generator families, evolution reduces motion JSD from 0.073 to 0.067, interaction JSD from 0.091 to 0.081, and OOS rate from 7.4\% to 6.0\%, bringing the generated behavior closer to the Waymo reference.}

% \vspace{-10pt}
\begin{table*}[!htbp]
\vspace{-10pt}
\centering
\scriptsize
\setlength{\tabcolsep}{8pt}
\renewcommand{\arraystretch}{1.0}
\caption{Independent post-evolution realism evaluation on held-out rollouts with maneuver-matched naturalistic references. Parent and Evolved report means over eight generator families.}
\label{tab:realism-validation}
\resizebox{0.7\linewidth}{!}{%
{%
\begin{tabular}{lccc}
\toprule
Source & Motion JSD$\downarrow$ & Interaction JSD$\downarrow$ & OOS rate$\downarrow$ \\
\midrule
Waymo split-half reference & 0.022 & 0.028 & 1.4\% \\
Parent, eight-family mean & 0.073 & 0.091 & 7.4\% \\
Evolved, eight-family mean & \textbf{0.067} & \textbf{0.081} & \textbf{6.0\%} \\
\bottomrule
\end{tabular}%
}%
}
\end{table*}

\vspace{-8pt}
\paragraph{{SafeBench.}}
{We next evaluate whether the observed performance generalizes to the scenario level. Table~\ref{tab:carla-main-generator-evolution} studies four scenario-level generators: AdvSim (AS)~\cite{wang2021advsim}, adversarial trajectory optimization (AT)~\cite{zhang2022adversarial}, ChatScene~\cite{zhang2024chatscene}, and a human-designed generator in CARLA~\cite{dosovitskiy2017carla}. 
% The CARLA attack score is the collision-weighted pressure $\mathrm{CR}(1-\mathrm{OS})$, so larger values indicate scenes that both induce collisions and reduce the benchmark driving score, while ARS measures adversarial-scene realism. 
Across all four sources, \method{} increases CR, lowers OS, and improves ARS after four evolution rounds. The significant PF-Area@3 gain further confirms a broadened attack-realism frontier.} {Appendix~\ref{app:supp-carla-family} reports the detailed results by method and evolution round.}
Figure~\ref{fig:carla-pareto-evolution} visualizes the cumulative local Pareto-frontier area for the four generators. Computed within family-route groups using ARS and attack severity, the mean search frontier gradually expands and broadens over evolutions.
Figure~\ref{fig:carla-cases} provides examples of original scenarios generated by ChatScene and evolved scenarios.

\begin{table*}[!htbp]
\vspace{-10pt}
\centering
\scriptsize
\setlength{\tabcolsep}{2pt}
\renewcommand{\arraystretch}{1.0}
\caption{SafeBench~\cite{xu2022safebench} generator evolution in CARLA. Attack is $\mathrm{CR}(1-\mathrm{OS})$; ARS is the adversarial realism score; {PF-Area@3 is the percentage gain in attack-ARS hypervolume over the parent frontier with up to three descendants, averaged over family-route groups.} Change reports relative gains for Attack and PF-Area@3, and point changes for ARS, CR, and OS.}
\label{tab:carla-main-generator-evolution}
\resizebox{0.9\linewidth}{!}{
\begin{tabular}{lccccccccccccc}
\toprule
 & \multicolumn{4}{c}{Parent} & \multicolumn{4}{c}{w/ \method} & \multicolumn{5}{c}{Change} \\
\cmidrule(lr){2-5}\cmidrule(lr){6-9}\cmidrule(l){10-14}
Method & Attack$\uparrow$ & ARS$\uparrow$ & CR$\uparrow$ & OS$\downarrow$ & Attack$\uparrow$ & ARS$\uparrow$ & CR$\uparrow$ & OS$\downarrow$ & Attack$\uparrow$ & ARS$\uparrow$ & CR$\uparrow$ & OS$\downarrow$ & PF-Area@3$\uparrow$ \\
\midrule
AS~\cite{wang2021advsim} & 0.260 & 0.544 & 0.631 & 0.589 & \cellcolor{green!10}0.382 & \cellcolor{green!10}0.615 & \cellcolor{green!10}0.791 & \cellcolor{green!10}0.517 & \cellcolor{green!12}+47.2\% & \cellcolor{green!12}+7.02pp & \cellcolor{green!12}+16.01pp & \cellcolor{green!12}-7.17pp & \cellcolor{green!12}+142.5\% \\
AT~\cite{zhang2022adversarial} & 0.197 & 0.514 & 0.537 & 0.633 & \cellcolor{green!10}0.330 & \cellcolor{green!10}0.607 & \cellcolor{green!10}0.729 & \cellcolor{green!10}0.548 & \cellcolor{green!12}+67.5\% & \cellcolor{green!12}+9.29pp & \cellcolor{green!12}+19.22pp & \cellcolor{green!12}-8.57pp & \cellcolor{green!12}+255.7\% \\
CS~\cite{zhang2024chatscene} & 0.394 & 0.729 & 0.812 & 0.516 & \cellcolor{green!10}0.507 & \cellcolor{green!10}0.924 & \cellcolor{green!10}0.934 & \cellcolor{green!10}0.457 & \cellcolor{green!12}+28.9\% & \cellcolor{green!12}+19.51pp & \cellcolor{green!12}+12.12pp & \cellcolor{green!12}-5.89pp & \cellcolor{green!12}+143.5\% \\
Human & 0.264 & 0.856 & 0.664 & 0.602 & \cellcolor{green!10}0.301 & \cellcolor{green!10}0.927 & \cellcolor{green!10}0.713 & \cellcolor{green!10}0.578 & \cellcolor{green!12}+13.8\% & \cellcolor{green!12}+7.13pp & \cellcolor{green!12}+4.89pp & \cellcolor{green!12}-2.41pp & \cellcolor{green!12}+73.7\% \\
\bottomrule
\end{tabular}
}
\end{table*}

\begin{figure}[H]
\centering
\includegraphics[width=0.8\linewidth]{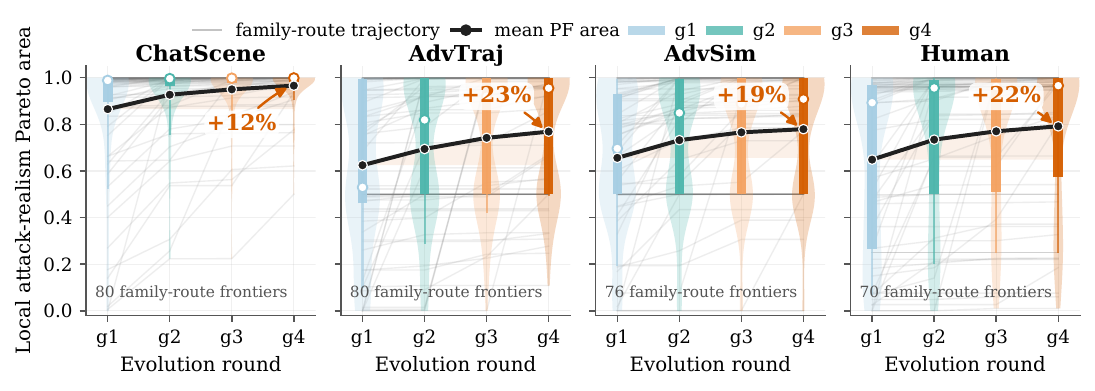}
\vspace{-14pt}
\caption{Pareto-frontier dynamics over evolution rounds. Gray curves track individual family-route groups, colored distributions indicate each round, and the black curve shows the mean Pareto area.}
\label{fig:carla-pareto-evolution}
\end{figure}

\begin{figure}[H]
\centering
\includegraphics[width=0.8\linewidth]{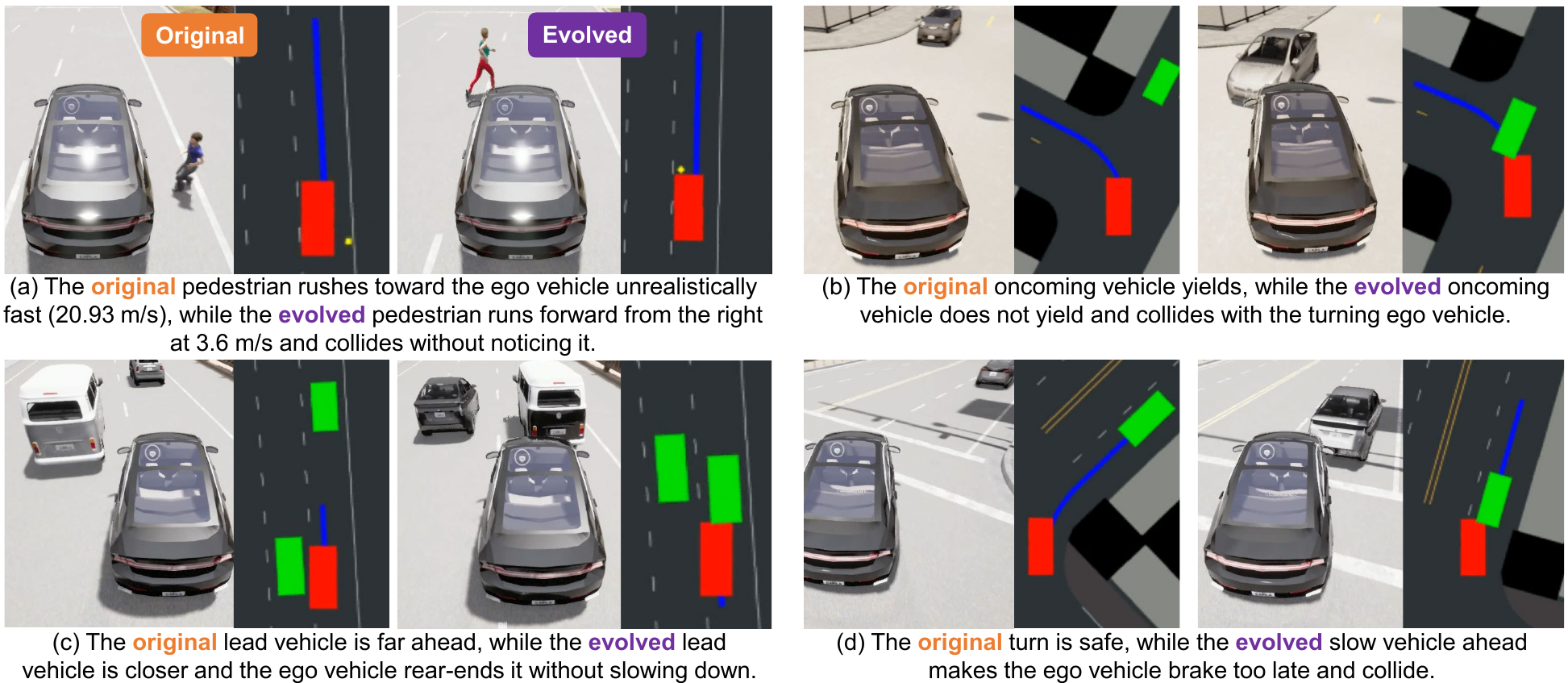}
\vspace{-10pt}
\caption{Qualitative examples of original and evolved scenarios generated by ChatScene.}
\label{fig:carla-cases}
\end{figure}

\subsection{Mechanism-Level Evolution Trajectories}
\label{sec:exp-mechanism-analysis}

% Keep this small figure and its explanation together in a fixed side-by-side row.
\noindent
\begin{minipage}[t]{0.39\linewidth}
\vspace{0pt}
\centering
\includegraphics[width=\linewidth]{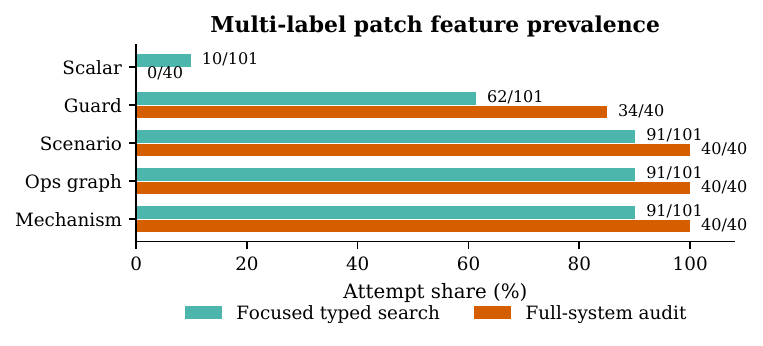}
\vspace{-16pt}
\captionof{figure}{Structured edit profiles.}
\label{fig:patch-surface-distribution}
\end{minipage}
\hfill
\begin{minipage}[t]{0.60\linewidth}
\vspace{0pt}
\paragraph{What the agents change.}
Figure~\ref{fig:patch-surface-distribution} reports the structure of evolved edits.
Most evaluated candidates modify mechanism modules, operator graphs, and scenario-conditioned policies, while scalar-only edits are rare.
This pattern indicates that the agents often discover reusable changes to how actors are selected, ranked, and composed into simulator interventions, rather than merely tuning parameters.
\end{minipage}
\par

\vspace{-5pt}
\paragraph{Evolution trajectory.}
Figure~\ref{fig:case-evolution-main} gives an example trace of how a generator changes over the course of evolution.
The score curve tracks the best candidate reached so far, while the bottom markers show proposals that were screened, revised by critics, or promoted into the generator lineage.
In this example, the agent first builds a time-to-conflict-based pressure mechanism, then refines fallback handling, timing, speed filters, and realism guards before reaching the final promoted generator.
The trajectory illustrates the intended role of the critic and evaluator: promising mechanisms are retained, while unsafe or unrealistic variants are filtered before they dominate the lineage. {Additional evolution traces are provided in Appendix~\ref{app:supp-evolution-traces}.}

\begin{figure}[!htbp]
\centering
\includegraphics[width=0.75\linewidth]{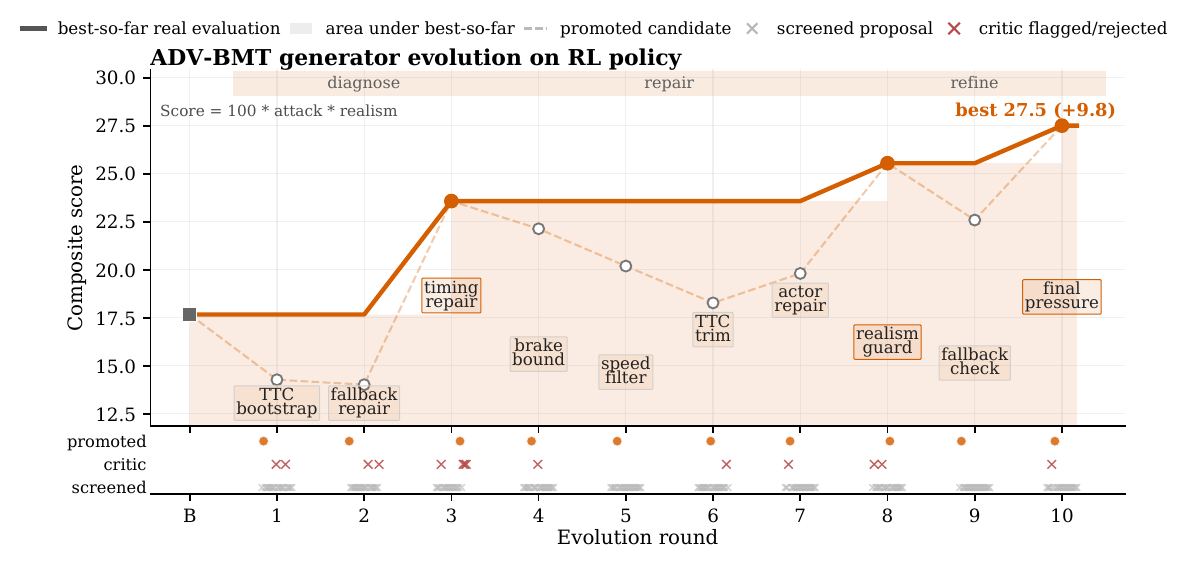}
\vspace{-13pt}
\caption{Evolution trajectory example. 
% The curve shows the best-so-far composite score; markers denote screened, critic-intervened, and promoted proposals.
}
\label{fig:case-evolution-main}
\end{figure}

\paragraph{{Component ablations.}}
{Table~\ref{tab:component-ablations} reports component ablations across three independent ADV-BMT evolution runs with a fixed RL ego policy. Full \method{} achieves the largest mean PF-Area@3 gain (31.0\%) and improves over the parent in all three runs. Removing the critic and repair lowers the number of frontier-improving candidates per 10 evaluations from 4.7 to 2.8 and increases RP from 1.78\% to 2.63\%. Removing long-term memory reduces PF-Area@3 to 19.1\%. The call-matched actor-only and non-LLM Pareto variants achieve 21.2\% and 10.6\%, respectively. These results support the contributions of review and repair, accumulated memory, and LLM-guided proposals.}

\begin{table*}[!htbp]
\centering
\scriptsize
\setlength{\tabcolsep}{5pt}
\renewcommand{\arraystretch}{1.0}
\caption{{Component ablations on ADV-BMT with the RL ego policy. FI/10 counts frontier-improving candidates per 10 evaluations; the actor-only control matches the total LLM calls.}}
\label{tab:component-ablations}
\resizebox{0.8\linewidth}{!}{%
{%
\begin{tabular}{lcccc}
\toprule
Variant & PF-Area@3 (\%)$\uparrow$ & Attack$\uparrow$ & RP (\%)$\downarrow$ & FI/10$\uparrow$ \\
\midrule
Full \method{} & \textbf{31.0\,\textpm\,4.2} & \textbf{0.238\,\textpm\,0.010} & \textbf{1.78\,\textpm\,0.12} & \textbf{4.7\,\textpm\,0.6} \\
w/o critic and repair & 26.8\,\textpm\,3.8 & 0.221\,\textpm\,0.011 & 2.63\,\textpm\,0.33 & 2.8\,\textpm\,0.6 \\
w/o long-term memory & 19.1\,\textpm\,3.5 & 0.218\,\textpm\,0.009 & 2.50\,\textpm\,0.22 & 3.8\,\textpm\,0.3 \\
Non-LLM Pareto search & 10.6\,\textpm\,2.8 & 0.205\,\textpm\,0.017 & 2.31\,\textpm\,0.25 & 2.2\,\textpm\,0.3 \\
Actor-only, calls matched & 21.2\,\textpm\,4.0 & 0.225\,\textpm\,0.012 & 2.17\,\textpm\,0.28 & 3.5\,\textpm\,0.4 \\
\bottomrule
\end{tabular}%
}%
}
\end{table*}

\vspace{-11pt}
\subsection{World Evaluator and Downstream Training}
\label{sec:exp-prioritization-training}

% {The complementary search diagnostics in Tables~\ref{tab:actor-critic-diagnostic} and~\ref{tab:kwidth-sensitivity} report proposal-side filtering and parallel search yield before archive selection.}
% Place the small evaluator table beside its corresponding discussion.
\noindent
\begin{minipage}[t]{0.56\linewidth}
\vspace{0pt}
\paragraph{World-guided candidate ranking.}
\looseness=-1
{Figure~\ref{fig:world-sidecar} shows the benefit of world-guided candidate ranking before simulator rollout. We further compare static and evolving evaluators in Table~\ref{tab:world-evaluator-evolution}. The static evaluator $W_0$ remains fixed, while the evolving evaluator $W_t$ updates using labels available from earlier rounds. Relative to $W_0$, $W_t$ needs 15\% fewer evaluations to reach the 90\% hypervolume target and raises top-10 yield by 10.8\%, with top-decile recall rising from 0.94 to 1.00.}
\end{minipage}
\hfill
\begin{minipage}[t]{0.42\linewidth}
\vspace{0pt}
\centering
\scriptsize
\setlength{\tabcolsep}{1pt}
\renewcommand{\arraystretch}{1.0}
\captionof{table}{{World evaluator evolution on the same candidate sequence and simulation budget. Evaluation count and top-10 yield are normalized to random ordering.}}
\label{tab:world-evaluator-evolution}
\resizebox{\linewidth}{!}{%
{%
\begin{tabular}{lccc}
\toprule
\shortstack[l]{Candidate\\ranking} & \shortstack{Norm. evals.\\for 90\% HV$\downarrow$} & \shortstack{Norm. top-10\\yield$\uparrow$} & \shortstack{Top-decile\\recall$\uparrow$} \\
\midrule
Random ordering & 1.00 & 1.00 & 0.72 \\
Static evaluator $W_0$ & 0.60\,\textpm\,0.04 & 1.30\,\textpm\,0.06 & 0.94 \\
Evolving evaluator $W_t$ & \textbf{0.51\,\textpm\,0.03} & \textbf{1.44\,\textpm\,0.05} & \textbf{1.00} \\
\bottomrule
\end{tabular}%
}%
}

\end{minipage}
\par

\vspace{-5pt}
\paragraph{Downstream policy training.}
We further evaluate whether evolved scenarios can improve downstream RL policies. 
In SafeBench, Table~\ref{tab:carla-chatscene-finetune} evaluates SAC policies fine-tuned with the original and evolved ChatScene scenes. 
Fine-tuning with evolved scenes reduces the average collision rate from $0.163$ to $0.106$ and improves the average overall score from $0.854$ to $0.884$. 
In MetaDrive, Figure~\ref{fig:sac-adv-bmt} shows SAC training curves on ADV-BMT scenarios before and after evolution, where the evolved scenarios yield lower adversarial crash rates and higher route completion. 
Thus, by balancing adversariality and realism, evolved generators effectively enhance downstream policy robustness.
% the evolved generators expose more valuable training cases, balancing adversariality and realism to improve downstream policy robustness.

\begin{table*}[!htbp]
\vspace{-6pt}
\centering
\scriptsize
\setlength{\tabcolsep}{2pt}
\renewcommand{\arraystretch}{1.0}
\caption{Downstream SAC fine-tuning performance. Original and Evolved denote RL policies fine-tuned with the original and evolved (w/ \method{}) ChatScene scenario, respectively.}
\label{tab:carla-chatscene-finetune}
\resizebox{0.85\linewidth}{!}{
\begin{tabular}{lcccccccccc}
\toprule
Metric & Algo. & Straight Obs. & Turn. Obs. & Lane Change & Veh. Pass. & Red-light & Unprot. Left & Right-turn & Crossing & Avg. \\
\midrule
CR$\downarrow$ & Original & 0.250 & 0.150 & {0.000} & \textbf{0.050} & 0.000 & 0.500 & 0.350 & 0.000 & 0.163 \\
 & Evolved & \textbf{0.100} & \textbf{0.050} & {0.000} & 0.200 & 0.000 & 0.500 & \textbf{0.000} & 0.000 & \cellcolor{green!12}\textbf{0.106} \\
\midrule
OS$\uparrow$ & Original & 0.800 & 0.847 & 0.949 & \textbf{0.921} & 0.948 & \textbf{0.711} & 0.705 & 0.949 & 0.854 \\
 & Evolved & \textbf{0.854} & \textbf{0.892} & \textbf{0.952} & 0.851 & \textbf{0.949} & 0.682 & \textbf{0.945} & \textbf{0.950} & \cellcolor{green!12}\textbf{0.884} \\
\bottomrule
\end{tabular}
}
\end{table*}
\vspace{-4pt}

\begin{figure}[!htbp]
\centering
\begin{minipage}[t]{0.54\linewidth}
\vspace{0pt}
\centering
\includegraphics[width=\linewidth]{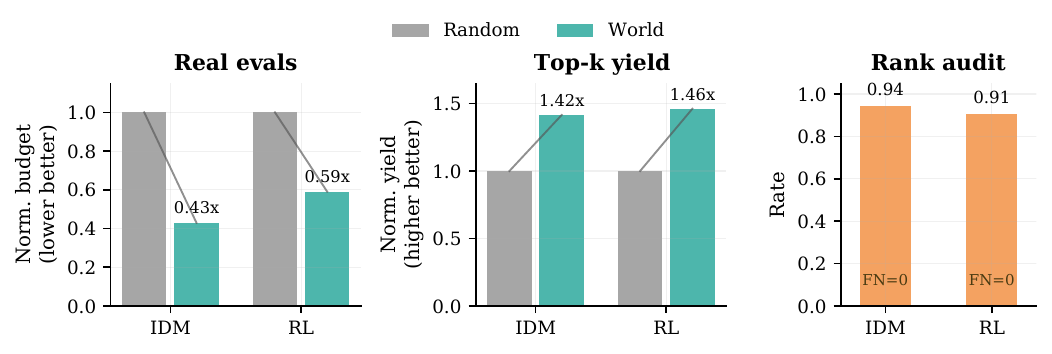}
\vspace{-20pt}
\caption{World-guided candidate ranking. We compare random ordering with world-ranked ordering for evaluation budget, top-$k$ yield, and ranking quality.}
\label{fig:world-sidecar}
\end{minipage}
\hfill
\begin{minipage}[t]{0.44\linewidth}
\vspace{0pt}
\centering
\includegraphics[width=\linewidth]{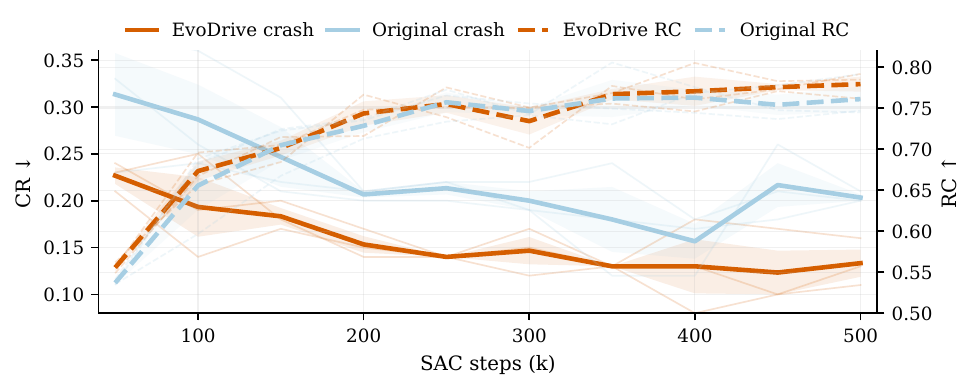}
\vspace{-20pt}
\caption{SAC training on ADV-BMT scenarios. CR is lower-is-better, and RC is higher-is-better, averaged over three seeds.}
\label{fig:sac-adv-bmt}
\end{minipage}
\end{figure}

%===============================================================================
\vspace{-10pt}
\section{Conclusion and Limitation}
\label{sec:conclusion}
% \vspace{-10pt}
In this paper, we presented \method{}, a fully automated agentic evolution framework for multi-objective adversarial scenario generation. By leveraging a specialized actor-critic architecture and a Pareto archive, \method{} iteratively improves scenario generation algorithms. Evaluations on benchmarks using MetaDrive and CARLA demonstrate that our framework yields structural improvements to existing generators and also effectively improves downstream policy robustness. 

\vspace{-8pt}
% \paragraph{Limitations.}
% \method{} focuses solely on evolving scenario generators against a fixed ego driving policy. Future work will address this by extending the framework to optimize driving algorithms directly, enabling an automated co-evolution of environments and policies.
% Due to the high cost and safety concerns associated with adversarial validation via real-world field testing, our results are limited to simulation platforms. Controlled hardware-in-the-loop validation remains an important direction for future work.
\paragraph{Limitations.}
\method{} currently assumes fixed ego policies, finite simulator budgets, and simulation-only validation, so it may miss failure modes tied to unseen policies, platforms, or real-world deployment. Due to the cost and safety risks of adversarial field testing, controlled hardware-in-the-loop validation and environment-policy co-evolution remain important future directions.

%===============================================================================

\clearpage
\endgroup
% The acknowledgments are automatically included only in the final and preprint versions of the paper.
% \acknowledgments{If a paper is accepted, the final camera-ready version will (and probably should) include acknowledgments. All acknowledgments go at the end of the paper, including thanks to reviewers who gave useful comments, to colleagues who contributed to the ideas, and to funding agencies and corporate sponsors that provided financial support.}

\acknowledgments{This research was in part supported by the National Natural Science Foundation of China (Project No. 524B2164, 52232015, and 52125208).}

%===============================================================================

\bibliography{references}  % .bib

\clearpage

\appendix
\section*{Appendix}
\startcontents[appendix]
\begingroup
\setcounter{tocdepth}{3}

\small                 % 控制目录字体大小，可换成 \small /
\linespread{0.88}\selectfont  % 控制目录行距，数值越小越紧凑
\setlength{\parskip}{3pt}     % 避免段间额外间距

\printcontents[appendix]{}{1}{\section*{Appendix Contents}\vskip-10pt\hrulefill\vskip-15pt}
\endgroup

\section{Related Work}\label{sec:related_work}

\paragraph{Safety-critical Autonomous Driving.}
Safety-critical scenario generation has been studied from multiple perspectives, including adaptive rare-event testing, adversarial perturbation of traffic participants, and data-driven generation under traffic priors. Early works such as \cite{feng2021nade,feng2023d2rl} treat the environment itself as an adaptive tester that exposes rare safety failures, while generator-based approaches \cite{hanselmann2022king,zhang2023cat,wang2021advsim,cao2022advdo} optimize behaviors of participants to produce failure-inducing interactions. Subsequent methods emphasize realism-preserving generation, e.g., by searching in learned manifolds or trajectory models~\cite{rempe2022generating,liu2026adv,mei2025llm,xu2025diffscene}. These generated scenarios are not only used for validation, but also for adversarial training and data augmentation of the driving policies \cite{hanselmann2022king,zhang2023cat,stoler2025seal,nie2025steerable,nie2026adv,zhang2024chatscene}.
Importantly, recent studies argue that useful generation should not over-optimize adversariality alone but instead balance realism, adversariality, and feasibility \cite{chen2024frea,stoler2025seal,nie2025steerable}. 
% FREA explicitly constrains generation with feasibility guidance to avoid inevitable crashes \cite{chen2024frea}; SEAL shows that simplistic objectives can produce overly aggressive or non-reactive adversaries \cite{stoler2025seal}; and SAGE enables test-time control over the trade-off rather than confining to a single fixed compromise during training \cite{nie2025steerable}. 
They move beyond pure failure maximization using feasibility constraints~\cite{chen2024frea}, realism-aware design~\cite{stoler2025seal}, and controllable trade-off mechanisms~\cite{nie2025steerable}, suggesting that the generation should be treated as a multi-objective problem rather than confined to a single fixed compromise.
However, existing methods often achieve this balance through fixed reward scalarizations, hand-crafted constraints, or other human-designed heuristics, which can be suboptimal and less generalizable across planners, datasets, and platforms.

\paragraph{LLM-based Agentic Discovery.}
Recent progress in LLM agents suggests a complementary route: instead of manually designing both a heuristic program and its search procedure, the search process itself can be optimized through executable feedback. Early agentic paradigms combine reasoning, tool use, self-reflection, memory, and skill accumulation to support long-horizon exploration and iterative improvement \cite{yao2022react,shinn2023reflexion,wang2023voyager}. More recent work extends this idea from task execution to autonomous discovery, where LLMs propose research artifacts, programs, or system configurations, evaluate them through executable tests or automated reviewers, and refine future proposals using archives, memories, or self-modifying code \cite{lu2024ai,yin2025godel,zhang2026darwin,novikov2025alphaevolve}. 
They show that evaluator-guided evolution over programs, harnesses, and even search strategies can yield sustained gains in scientific discovery and systems optimization~\cite{lange2025shinkaevolve,lee2026meta,liu2026evox,qu2026coral}.
Across these studies, a common pattern is to replace a fixed hand-designed search policy with an adaptive outer loop that learns from prior candidates and evaluator feedback. However, most existing agentic discovery systems are evaluated on coding, mathematics, research automation, or workflow-optimization tasks, rather than simulator-grounded safety validation for autonomous driving. Consequently, it remains unclear how such self-improving search should be constrained and adapted to discover scenario generators that must jointly satisfy realism, adversariality, and feasibility under expensive closed-loop evaluation.

\section{Extended Method and Implementation Details}
\label{app:extended-method}

This appendix complements Section~\ref{sec:method} by specifying the implementation-level details that are used by the method described in the main text. It focuses on the state carried across epochs, the validation and compilation contract, the conversion from rollout diagnostics to archive statistics, and the memory signals used to condition later proposals.

\subsection{Adversarial Scenario Generation}
\label{app:adv-generation-variables}

For a scene $s$, the scene-level decision variable in Eq.~\eqref{eq:prelim-generator-factorization} can be written conceptually as
\begin{equation}
    \omega={(u,i,\iota,\vartheta,\chi)},
    \label{eq:app-scene-level-omega}
\end{equation}
where $u$ denotes a scene mode or interaction cluster, $i$ indexes the non-ego actor or actor set to be modified, $\iota$ specifies the event-aligned interaction window, $\vartheta$ contains bounded motion or behavior parameters, and $\chi$ collects realism and diversity guards. A generator program induces these decisions through the chain
\begin{equation}
\begin{aligned}
q_c(\omega\mid\psi(s))
=&\;q_c(u\mid\psi(s))\,
q_c(i\mid u,\psi(s))\,
{q_c(\iota\mid i,u,\psi(s))} \\
&\;\times {q_c(\vartheta\mid \iota,i,u,\psi(s))\,
q_c(\chi\mid \vartheta,\iota,i,u,\psi(s)).}
\end{aligned}
\label{eq:app-generator-decision-factorization}
\end{equation}
The factorization is conceptual: a practical generator may implement some factors by deterministic ranking, top-$k$ search, or bounded grid search, which corresponds to a degenerate distribution over admissible decisions. {The key implementation point is that all factors are evaluated before closed-loop rollout and must produce decisions that satisfy the family interface and scene-specific feasibility constraints.} Generator-level evolution changes the scoring rules, mechanisms, or guards that define these factors. Scenario-level specialization then selects a concrete $\omega$ for each seed and compiles it into $\zeta=\compiler(c,\omega;s)$ before closed-loop rollout.

\subsection{Full Algorithmic Flow}
\label{app:algorithms}
\begin{algorithm}[!htbp]
\caption{\method{} generator-program evolution loop}
\label{alg:evodrive_loop}
\algcompact
\begin{algorithmic}[1]
\Require Fixed ego policy $\egopolicy$; seed pool $\seedset$; adapter registry {$\mathfrak{I}=\{\mathcal{I}_b\}_b$}; initial lineage store $\mathcal{L}_0$; archive $\mathcal{P}_0$; optional world snapshot $W_{\phi_0}$; budget $T$
\Ensure Pareto archive $\mathcal{P}_T$, lineage store $\mathcal{L}_T$, memory {$\mathcal{Q}_T$}, evidence bundle $\mathcal{B}_T$
\State Initialize attempts $\mathcal{E}_0\gets\emptyset$ and memory {$\mathcal{Q}_0\gets\emptyset$}.
\For{$t=0$ to $T-1$}
    \State Select lineage $\ell_t$ and active parent program $p_t\gets\mathcal{L}_t(\ell_t)$.
    \State Retrieve causal context {$\mathcal{K}_t^{\ell}\gets\Call{RetrieveMemory}{\mathcal{Q}_t,p_t,t}$}.
    \State $\delta_t \gets \Call{ActorCriticMutation}{p_t,{\mathcal{K}_t^{\ell}},\mathfrak{I}}$.
    \If{{$\delta_t=\bot$}}
        \State {Record critic-rejection attempt $a_t$; $\mathcal{E}_{t+1}\gets\mathcal{E}_t\cup\{a_t\}$; $\mathcal{Q}_{t+1}\gets\Call{CurateMemory}{\mathcal{Q}_t,a_t}$.}
        \State $\mathcal{P}_{t+1}\gets\mathcal{P}_t$; $\mathcal{L}_{t+1}\gets\mathcal{L}_t$; $W_{\phi_{t+1}}\gets W_{\phi_t}$.
        \State \textbf{continue}
    \EndIf
    \State $p'_t \gets \Call{ApplyPatch}{p_t,\delta_t}$.
    \If{$\Call{ValidateProgram}{p'_t,\delta_t,\mathfrak{I}}=0$}
        \State {Record contract-violation attempt $a_t$; $\mathcal{E}_{t+1}\gets\mathcal{E}_t\cup\{a_t\}$; $\mathcal{Q}_{t+1}\gets\Call{CurateMemory}{\mathcal{Q}_t,a_t}$.}
        \State $\mathcal{P}_{t+1}\gets\mathcal{P}_t$; $\mathcal{L}_{t+1}\gets\mathcal{L}_t$; $W_{\phi_{t+1}}\gets W_{\phi_t}$.
        \State \textbf{continue}
    \EndIf
    \State $c_t \gets \Call{CompileCandidate}{p'_t}$.
    \If{$W_{\phi_t}$ is enabled}
        \State $d_t \gets \Call{WorldRoute}{W_{\phi_t},c_t,{\mathcal{S}^{\mathrm{route}}_t}}$.
    \Else
        \State $d_t \gets \mathrm{select}$.
    \EndIf
    \If{$d_t=\mathrm{reject}$}
        \State {Record routed-rejection attempt $a_t$; $\mathcal{E}_{t+1}\gets\mathcal{E}_t\cup\{a_t\}$; $\mathcal{Q}_{t+1}\gets\Call{CurateMemory}{\mathcal{Q}_t,a_t}$.}
        \State $\mathcal{P}_{t+1}\gets\mathcal{P}_t$; $\mathcal{L}_{t+1}\gets\mathcal{L}_t$; $W_{\phi_{t+1}}\gets W_{\phi_t}$.
        \State \textbf{continue}
    \EndIf
    \State $Y_t \gets \Call{RealSimulatorEvaluate}{\simulator,\egopolicy,c_t,\mathcal{S}_{\mathrm{eval}}}$.
    \State {$a_t$} $\gets \Call{BuildAttemptRecord}{c_t,p_t,p'_t,d_t,Y_t}$.
    \State $(\mathrm{accept},\mathcal{P}_{t+1}) \gets \Call{ParetoAccept}{\mathcal{P}_t,{a_t}}$.
    \State $\mathcal{L}_{t+1}\gets\Call{UpdateLineage}{\mathcal{L}_t,p_t,p'_t,{a_t},\mathrm{accept}}$.
    \State {$\mathcal{Q}_{t+1}\gets\Call{CurateMemory}{\mathcal{Q}_t,a_t}$}; $\mathcal{E}_{t+1}\gets\mathcal{E}_t\cup\{{a_t}\}$.
    \If{$\Call{WorldUpdateReady}{W_{\phi_t},\mathcal{E}_{t+1}}$}
        \State $W_{\phi_{t+1}}\gets\Call{WorldAgentUpdate}{W_{\phi_t},\mathcal{E}_{t+1}}$.
    \Else
        \State $W_{\phi_{t+1}}\gets W_{\phi_t}$.
    \EndIf
\EndFor
\State \Return $\mathcal{P}_T,\mathcal{L}_T,{\mathcal{Q}_T},\Call{BuildEvidenceBundle}{\mathcal{E}_T,\mathcal{P}_T,\mathcal{L}_T,{\mathcal{Q}_T}}$.
\end{algorithmic}
\end{algorithm}

\begin{algorithm}[!htbp]
\caption{Actor-critic program mutation}
\label{alg:actor_critic_mutation}
\algcompact
\begin{algorithmic}[1]
\Require Parent solution program $p$; memory context {$\mathcal{K}$}; family interface {$\mathcal{I}_b$}; maximum repair rounds $K_r$
\Ensure Bounded patch $\delta$ or rejection token $\bot$
\State $\delta^{(0)}\gets \Call{ProgramActor}{p,{\mathcal{K}},{\mathcal{I}_b}}$.
\For{$k=0$ to $K_r$}
    \State $r_s\gets\Call{StructuralCritic}{p,\delta^{(k)},{\mathcal{I}_b}}$.
    \State $r_a\gets\Call{AttackCritic}{p,\delta^{(k)},{\mathcal{K}}}$.
    \State $r_r\gets\Call{RealismCritic}{p,\delta^{(k)},{\mathcal{K}}}$.
    \State $r_i\gets\Call{IsolationCritic}{p,\delta^{(k)},{\mathcal{I}_b}}$.
    \If{$r_s,r_a,r_r,r_i$ all approve}
        \State \Return $\delta^{(k)}$.
    \ElsIf{any critic rejects without repair}
        \State \Return $\bot$.
    \Else
        \State $\delta^{(k+1)}\gets\Call{ActorRepair}{p,\delta^{(k)},r_s,r_a,r_r,r_i}$.
    \EndIf
\EndFor
\State \Return $\bot$.
\end{algorithmic}
\end{algorithm}

\begin{algorithm}[!htbp]
\caption{World agent evaluator self-evolution}
\label{alg:world_agent_update}
\algcompact
\begin{algorithmic}[1]
\Require Active snapshot $W_\phi$; real-label attempts $\mathcal{E}^{\mathrm{real}}$; prescreen records $\mathcal{E}^{W}$; allowed update surface $\mathcal{U}_W$
\Ensure Promoted snapshot $W_{\phi'}$ or original snapshot $W_\phi$
\State $\mathcal{D}^{W}\gets\Call{BuildCalibrationSet}{\mathcal{E}^{\mathrm{real}},\mathcal{E}^{W}}$.
\State $a_W\gets\Call{AuditWorld}{W_\phi,\mathcal{D}^{W}}$.
\If{$\Call{SufficientEvidence}{a_W}=0$}
    \State \Return $W_\phi$.
\EndIf
\State $\mathcal{T}_W\gets\Call{BuildWorldAgentTask}{W_\phi,a_W,\mathcal{U}_W}$.
\State $\Delta\phi\gets\Call{WorldAgent}{\mathcal{T}_W}$.
\If{$\Delta\phi\notin\mathcal{U}_W$}
    \State \Return $W_\phi$.
\EndIf
\State $W_{\tilde{\phi}}\gets\Call{MaterializeSnapshot}{W_\phi,\Delta\phi}$.
\State $j\gets\Call{WorldJudge}{W_{\tilde{\phi}},W_\phi,\mathcal{D}^{W},a_W}$.
\If{$j=\mathrm{accept}$}
    \State {Record promoted snapshot and later-epoch usage requirement.}
    \State \Return $W_{\tilde{\phi}}$.
\Else
    \State Record rejected evaluator update.
    \State \Return $W_\phi$.
\EndIf
\end{algorithmic}
\end{algorithm}

\begin{algorithm}[!htbp]
\caption{Parallel candidate epoch with serial acceptance}
\label{alg:parallel_epoch}
\algcompact
\begin{algorithmic}[1]
\Require Immutable epoch snapshot {$z_t=(p_t,h_t,\ell_t,\seedset_t,W_{\phi_t},\mathcal{K}_t^\ell,\upsilon_t)$}; number of workers $K$
\Ensure Deterministically accepted state update
\For{$k=1$ to $K$ in parallel}
    \State Worker $k$ receives $z_t$ and proposes candidate record {$\tilde{c}_{t,k}$}.
    \State Worker $k$ may write local proof, patch, validation trace, and candidate-local records.
    \State Worker $k$ is not allowed to mutate archive, active parent, lineage store, or shared memory.
\EndFor
\State Wait until all workers reach the epoch barrier.
\State Sort candidate records by deterministic key.
\For{candidate record {$\tilde{c}_{t,k}$} in sorted order}
    \State Apply validation, routing, real evaluation when selected, and archive acceptance.
\EndFor
\State Update archive, lineage, and memory exactly once through the serial acceptor.
\end{algorithmic}
\end{algorithm}

\subsection{Program, Candidate, and Evidence Objects}
\label{app:program-objects}

Table~\ref{tab:app-program-schema} summarizes the conceptual generator program fields that correspond to Eq.~\eqref{eq:method-program-state}.

\begin{table}[!htbp]
\centering
\small
\caption{Conceptual generator program schema.}
\label{tab:app-program-schema}
\begin{tabularx}{\linewidth}{>{\ttfamily}lX}
\toprule
Field & Role \\
\midrule
solution\_id, lineage\_id & Identify the node and the lineage to which it belongs. \\
source\_family & Generator family or seed baseline that defines the legal interface. \\
generation & Evolution depth from the initial generator. \\
operator\_graph & Typed graph of simulator operators such as selection, timing, motion, filtering, and diversity. \\
parameters & Bounded numeric or categorical generator settings. \\
scenario\_conditioned\_policy & Rule that maps seed features to operator settings. \\
mechanism\_modules & Reusable declarative mechanisms inserted into supported operators. \\
compile\_contract & Simulator backend, required constraints, and compile target. \\
program\_hash & Stable hash of normalized program content. \\
ancestor\_chain & Parent ids, child ids, and edit hashes used for audit. \\
\bottomrule
\end{tabularx}
\end{table}

A lineage store attaches every proposal attempt $a_t$ to its source lineage and parent state. If a child program is accepted, the lineage store promotes the child as the active solution for that lineage. If the child is rejected, invalid, or routed away from real evaluation, the corresponding attempt remains recorded as failed-proposal evidence rather than being discarded. Thus, the generator history is represented as an auditable sequence
$p_0 \xrightarrow{\delta_0} p_1 \xrightarrow{\delta_1} \cdots \xrightarrow{\delta_{t-1}} p_t$,
where each proposed transition is either validated and promoted or retained as negative evidence for later memory curation.

A local edit contains the parent identity, parent program hash, source family, a list of typed edit operations, and a short rationale. Legal operations are limited to add, replace, and remove under the paths in Section~\ref{subsec:method-generator-programs}. 
A candidate is the simulator-based description obtained after applying and compiling an edit. An attempt record $a_t$ attaches the candidate to its evidence trail: parent lineage, edit metadata, validation result, compile output, world routing record, simulator metrics when available, archive decision, seed-level outcomes, and proposal provenance. Memory curation operates over attempt records rather than raw logs, which keeps retrieval compact and semantically typed.

\subsection{Validation Rules and Compilation}
\label{app:validation-compilation}

The deterministic validator is decomposed into independent gates. For Appendix notation, we write
\begin{equation}
\begin{aligned}
    &V_{\mathrm{app}}(p'_t,\delta_t)=
    \prod_{j\in\mathcal{J}_{\mathrm{app}}}V_j(p'_t,\delta_t),
    \\
    &\mathcal{J}_{\mathrm{app}}=
    \{\mathrm{schema},\mathrm{parent},\mathrm{interface},\mathrm{isolation},\mathrm{protected},\mathrm{mechanism},\mathrm{compile}\}.
\label{eq:app-validation-gates}
\end{aligned}
\end{equation}

The schema gate checks required fields and type constraints. The parent gate checks parent identity, parent hash, source family, and generation. The interface gate checks that every edit path belongs to the allowed generator surface. The isolation gate prevents a lineage from importing family-specific mechanisms from another source family. The protected-resource gate prevents edits to the ego policy, benchmark manifest, simulator evaluator, locked seed split, and archive acceptor. The mechanism gate rejects unsupported side effects and unsupported declarative expressions. The compile gate verifies that the resulting program can be lowered to simulator operators. The validator is fail-closed: any failed gate produces a recorded invalid attempt and stops the real simulation.

The compiler translates a valid generator into an operator composition. A typical scene-level compilation has
\begin{equation}
{
    \zeta(c,s)=\compiler_b(c,\omega;s)=
    \left(
    O_{\mathrm{select}},
    O_{\mathrm{actor}},
    O_{\mathrm{timing}},
    O_{\mathrm{motion}},
    O_{\mathrm{search}},
    O_{\mathrm{filter}},
    O_{\mathrm{div}}
    \right).
}
\label{eq:app-compiler}
\end{equation}
{$O_{\mathrm{select}}$ selects applicable seeds or scene clusters.} $O_{\mathrm{actor}}$ ranks non-ego actors that can be modified. {$O_{\mathrm{timing}}$ chooses interaction windows such as cut-in, merge, crossing, or braking intervals.} $O_{\mathrm{motion}}$ instantiates trajectory changes under bounded dynamics. {$O_{\mathrm{search}}$ searches over legal parameter ranges. $O_{\mathrm{filter}}$ removes map-inconsistent or dynamically infeasible outcomes.} $O_{\mathrm{div}}$ discourages collapse to repeated templates. The compiler stores source-family provenance, operator types, mechanism tags, {compiled metadata}, and simulator-backend metadata in the attempt record.

\subsection{Objective Computation and Archive Statistics}
\label{app:objective-computation}

\paragraph{Rollout-level quantities.}
For each evaluated scenario seed $s$, the simulator returns a rollout $\tau(c,s)$ with event indicators and geometric diagnostics. The attack score $a(\tau)\in[0,1]$ is the target-failure indicator or severity score used by the benchmark. The realism score $r(\tau)\in[0,1]$ aggregates feasibility checks such as map compliance, collision validity, route completion, and bounded kinematics. The candidate-level objectives are the empirical means over the evaluation subset, as in Eq.~\eqref{eq:prelim-objectives}. {Equivalently, these quantities are finite-sample estimates of the attack and realism utilities under $p_c(s,\omega,\zeta,\tau,y)$ in Eq.~\eqref{eq:prelim-generator-factorization}.}

\paragraph{Realism penalty.}
Tables~\ref{tab:metadrive-main-generator-evolution} and \ref{tab:metadrive-paper-baseline-profile-detail} report the realism penalty
\begin{equation}
    \mathrm{RP}(c)=100\cdot(1-R(c)),
    \label{eq:app-realism-penalty}
\end{equation}
so that lower values indicate more realistic scenes. For the MetaDrive experiments, $R(c)$ is the candidate-level average of rollout realism diagnostics:
\begin{equation}
\begin{aligned}
    R_{\mathrm{MD}}(c)
    &=
    \frac{1}{|\seedset_{\mathrm{eval}}|}
    \sum_{s\in\seedset_{\mathrm{eval}}}
    r_{\mathrm{MD}}\bigl(\tau(c,s)\bigr),
    \\
    r_{\mathrm{MD}}(\tau)
    &=
    \sum_{j\in\mathcal{J}_{\mathrm{MD}}}w_j d_j(\tau),
    \quad \sum_j w_j=1,\; w_j\ge 0.
\end{aligned}
\label{eq:app-metadrive-realism-score}
\end{equation}
Here $\mathcal{J}_{\mathrm{MD}}$ includes the MetaDrive feasibility diagnostics used by the evaluator, including map and lane validity, initial non-overlap, collision validity, route progress, trajectory continuity, and bounded acceleration or jerk. Binary diagnostics use $d_j(\tau)=\mathbf{1}[\text{check }j\text{ is satisfied}]$, while soft violations use $d_j(\tau)=1-\min\{v_j(\tau)/\rho_j,1\}$ with non-negative violation $v_j$ and tolerance scale $\rho_j$.
% This is only a display transform of $R(c)$; archive acceptance still uses the larger-is-better objective $R(c)$.

\paragraph{Matched-budget frontier area.}
For a method family $g$, let $\mathcal{P}^{0}_g$ be the parent frontier and let $\mathcal{P}^{k}_g$ be the non-dominated set formed after adding {up to $k$ eligible descendants preselected under the archive rule using evolution results and frozen before held-out evaluation}. The reported matched-budget frontier statistic is
\begin{equation}
    \mathrm{PF\mbox{-}Area@}k(g)
    =
    100\cdot
    \frac{\mathrm{HV}(\mathcal{P}^{k}_g)-\mathrm{HV}(\mathcal{P}^{0}_g)}
    {\max\{\mathrm{HV}(\mathcal{P}^{0}_g),\varepsilon\}},
    \label{eq:app-pf-area}
\end{equation}
where $\mathrm{HV}$ is computed in the attack-realism plane with a fixed reference point and $\varepsilon$ prevents division by zero for degenerate parent frontiers. In the paper tables, $k=3$ gives the same reporting budget across generator families.

\subsection{Proposal and Repair}
\label{app:agent-roles}

{The proposal actor receives the parent program, generator interface, current search target, and retrieved memory context $\mathcal{K}_t^\ell$.} It returns one local edit and a rationale. The rationale is used by critics and memory curation, but it has no authority over acceptance unless the edit compiles and receives simulator labels.

The structural critic checks syntax, required fields, edit paths, parent hash, and mechanism declarations. The attack critic checks whether the edit plausibly increases failure exposure of the fixed ego policy, e.g., by sharpening time-to-conflict pressure, improving actor ranking, expanding vulnerable seed coverage, or improving timing search. The realism critic checks lane validity, map consistency, initial overlap, acceleration, jerk, trajectory continuity, route compatibility, and controller feasibility. The isolation critic checks that the edit stays within the active family interface.

If the critics request repair, the {repair agent} may reduce parameter magnitudes, add realism guards, replace unsupported mechanisms with supported ones, or move an operation to a legal edit path. Repair is bound by a small budget. A repaired edit still passes through deterministic validation before compilation. Thus, critic approval is helpful but never sufficient for simulation or archive promotion.

\subsection{World Evaluator and Evaluator Evolution}
\label{app:world-details}

The world evaluator uses a feature map $x(c,s)$ with five groups of signals: generator-family indicators, local-edit summaries, operator-graph statistics, seed-level geometric descriptors, and lightweight interaction estimates. Interaction estimates may include relative distance, relative speed, time to conflict, route deviation, and realism diagnostics computed without a full closed-loop rollout.

For each candidate-seed pair, the evaluator predicts attack, realism, and uncertainty. Candidate-level predictions are obtained by aggregating over the routed seed subset. {Uncertainty aggregation uses a conservative statistic, such as a maximum or upper quantile, so that sparse or unstable regions remain eligible for audit.} Calibration mixes model predictions with empirical statistics from simulator-labeled records. A simple form is
\begin{equation}
    \widehat{A}_{\mathrm{cal}}
    = \beta_A \widehat{A}_{\mathrm{raw}} + (1-\beta_A)\bar{A}_{g},
    \qquad
    \widehat{R}_{\mathrm{cal}}
    = \beta_R \widehat{R}_{\mathrm{raw}} + (1-\beta_R)\bar{R}_{g},
    \label{eq:app-world-calibration}
\end{equation}
where $\bar{A}_{g}$ and $\bar{R}_{g}$ are group-level statistics for a family, seed cluster, or mechanism class, with fallback to global statistics when evidence is sparse. {The weights $\beta_A$ and $\beta_R$ are calibration weights and are distinct from the routing thresholds $\lambda_A,\lambda_R,\lambda_U$ in Eq.~\eqref{eq:method-routing}.}

The routing policy compares calibrated predictions and uncertainty to thresholds. A candidate is selected when predicted attack and realism are high enough and uncertainty is low enough. It is audited when uncertainty is high, when a false-rejection monitor requests labels, or when the candidate belongs to a sparse calibration region. It is rejected before simulation only when the evaluator is sufficiently confident that the candidate is of low value or invalid-looking. Rejected candidates are not archive entries; they are routing records used for calibration and later analysis.

Evaluator self-evolution is restricted to the evaluator configuration. The world Agent may change feature choices, calibration policy, ranking policy, uncertainty policy, or routing thresholds. {The allowed update surface excludes scenario-generator edits, ego-policy edits, simulator-evaluator edits, and changes to simulator labels.} A candidate evaluator snapshot is promoted only if the world judge verifies feasibility, protected-resource compliance, ranking quality, realism calibration, and false-negative behavior under audit evidence. Rejected snapshots remain in the evaluator lineage as negative evidence.

\subsection{Archive, Memory, and Retrieval}
\label{app:archive-memory}

An archive entry stores candidate identity, source family, ego-policy identity, generator lineage, attack metrics, realism metrics, diagnostics, evaluation seed subset, simulator version, {compiled-operator metadata}, and acceptance gate. The archive analyzer extracts the non-dominated frontier, computes frontier coverage or hypervolume when needed, and groups entries by family, seed cluster, and mechanism tags.

% Memory cards are normalized summaries derived from attempt records. Failure memories describe invalid edits, failed compilation, low-realism mechanisms, low-attack mechanisms, and world false rejections. Success motifs describe accepted mechanisms and the conditions under which they were effective. Critic notes preserve concise structural, attack, realism, and isolation feedback. Lineage memories summarize progress for each source family. Reflection plans suggest whether the next epoch should repair, reuse, diversify, or pivot.

Let $a_t$ denote the attempt record completed at epoch $t$. A curator $C$ updates the structured memory set as
\begin{equation}
    \mathcal{Q}_{t+1}=\mathcal{Q}_t\cup C(a_t),
\end{equation}
where $C(a_t)$ emits normalized memory cards rather than raw logs. Failure memories describe invalid edits, failed compilation, low-realism mechanisms, low-attack mechanisms, and world false rejections. Success motifs describe accepted mechanisms and the conditions under which they were effective. Critic notes preserve concise structural, attack, realism, and isolation feedback. Lineage memories summarize progress for each source family. Reflection plans suggest whether the next epoch should repair, reuse, diversify, or pivot.

For a proposal to be generated at epoch $t+1$, retrieval first applies the causal eligibility set
\begin{equation}
    \mathcal{Q}_{\le t}^{\mathrm{elig}}=\{u\in\mathcal{Q}_{\le t}: \mathrm{epoch}(u)\le t\}.
\end{equation}

A typical retrieval score for memory card $u\in{\mathcal{Q}_{\le t}^{\mathrm{elig}}}$ and parent program $p_t$ is
\begin{equation}
\begin{aligned}
    \mathrm{score}(u,p_t)=&\;\alpha_1\mathbf{1}[\ell(u)=\ell(p_t)]
    +\alpha_2\,\mathrm{sim}_{\mathrm{family}}(u,p_t)
    +\alpha_3\,\mathrm{sim}_{\mathrm{mechanism}}(u,p_t) \\
    &+\alpha_4\,\mathrm{sim}_{\mathrm{scene}}(u,p_t)
    +\alpha_5\,\mathrm{recency}(u)
    -\alpha_6\mathbf{1}[\mathrm{future}(u,t)].
\end{aligned}
\label{eq:app-memory-score}
\end{equation}
{The retrieved context is then $\mathcal{K}_{t+1}^\ell=\mathrm{TopM}_{u\in\mathcal{Q}_{\le t}^{\mathrm{elig}}}\,\mathrm{score}(u,p_t)$.} Memory created after epoch $t$ cannot be retrieved for a proposal at epoch $t+1$. Shared memory from other lineages is allowed only when it describes abstract mechanisms and contains no family-specific implementation payload or benchmark-score shortcut.

\subsection{Parallel Candidate Epochs}
\label{app:parallel-epochs}

A parallel epoch begins by freezing an immutable snapshot
\begin{equation}
{
    z_t=(p_t,h_t,\ell_t,\seedset_t,W_{\phi_t},\mathcal{K}_t^\ell,\upsilon_t),
}
\end{equation}
where $p_t$ is the parent generator, $h_t$ is its hash, $\ell_t$ is the lineage id, $\seedset_t$ is the seed subset, $W_{\phi_t}$ is the active World snapshot, {$\mathcal{K}_t^\ell$ is the retrieved memory context, and $\upsilon_t$ is the parallel resource policy.} Each worker receives $z_t$ and can write candidate-local records such as the proposal, local edit, critic reports, validation trace, and routing prediction. Workers cannot update shared lineage, archive, memory, or active parent state.

After all workers reach the epoch barrier, a serial acceptor sorts candidate artifacts by a deterministic key and applies the standard validation, routing, simulation, archive, and memory updates. This makes parallel generation equivalent to a deterministic serial state transition, while still using multiple proposal samples to increase search diversity.
Consequently, parallelism changes only the number of proposal samples available at an epoch, not the logical order in which archive, lineage, and memory states are mutated.

% \newpage
\section{Detailed Experimental Setups}
\label{app:experimental-setups}

This section provides the evaluation protocols for the experiment results reported in Section~\ref{sec:experiment} and Appendix~\ref{app:supplementary-results}. Details include the benchmark, metric definitions, baseline implementation, and evaluation procedures needed to interpret the reported results.

\subsection{MetaDrive Protocol}
\label{app:metadrive-setup}

\paragraph{Benchmark and ego policies.}
Our study in MetaDrive~\cite{li2022metadrive} simulator evaluates adversarial scenario generators against fixed ego policies. Following \citet{zhang2023cat}, we use one rule-based IDM policy and one learned RL policy to cover both reactive and learned closed-loop behavior. During generator evolution, the ego policy remains fixed. There is no policy training performed inside the evolution loop. Candidate generators are developed on a rotating subset drawn from a fixed scenario evolution pool and are then reported on a protected held-out evaluation suite with 100 scenarios. {The scenario pool is initialized from Waymo Open Motion Dataset episodes~\cite{ettinger2021large}, preserving the map, route, and initial traffic state when replayed in MetaDrive.}
We preprocessed a total of 500 scenarios for the study.

\paragraph{Generator families.}
Due to the closed-loop interaction and real-world traffic scenarios, MetaDrive is widely used to study safety-critical scenario generation and adversarial training \cite{zhang2023cat,stoler2025seal,mei2025llm,nie2025steerable,nie2026adv,liu2026adv}.
Following this line of work, our experiments cover model-based and optimization-based adversarial generators, including CAT \cite{zhang2023cat}, ADV-BMT \cite{liu2026adv}, AT \cite{zhang2022adversarial}, SAGE \cite{nie2025steerable}, LLM-Attacker \cite{mei2025llm}, KING \cite{hanselmann2022king}, SEAL \cite{stoler2025seal}, and STRIVE \cite{rempe2022generating}. 
These generators propose trajectory perturbation of prescribed or selected adversarial agents to induce collision with the ego vehicle. Note that CAT, ADV-BMT, SEAL, LLM-Attacker, and SAGE are originally designed for MetaDrive, while other baselines are properly adapted to the same setting.
Each baseline is represented through the same generator interface: a parent generator defines the editable surface, valid descendants are compiled into simulator operators in MetaDrive, and accepted descendants enter the same Pareto archive. 

\paragraph{Evolution protocol.}
For the MetaDrive experiments, each lineage is evolved for 10 epochs with 20 parallel proposal slots per epoch. Each epoch uses a 20-scenario rotating development window drawn from the fixed 240-scenario evolution pool. It then generates bounded candidate edits from the active parent, reviews them through the actor-critic proposal loop, validates and compiles the surviving edits, routes selected candidates to real simulator evaluation, and updates the archive only from simulator-derived labels. The reporting table compares the parent generator, the selected active generator after evolution, and the matched-budget frontier area obtained from eligible descendants. {The same protocol is used across generator families where the ego policy is fixed, the admissible edit surface is family-specific but declared before evolution, and held-out reporting is performed only after candidate generation and archive selection are completed.}

\paragraph{Metrics.}
Attack is the target-failure (collision) rate on 100 held-out test scenarios, where larger values indicate more adversarial scenarios. Realism is reported as $\mathrm{RP}=100(1-R)$, where lower values indicate fewer realism penalties. 
{Thus Tables~\ref{tab:metadrive-main-generator-evolution} and \ref{tab:metadrive-paper-baseline-profile-detail} display the same objectives as Eq.~\eqref{eq:prelim-objectives}, with Attack corresponding to $A(c)$ and RP serving as the monotone transform of the MetaDrive realism score in Eq.~\eqref{eq:app-metadrive-realism-score}.}
PF-Area@3 is the matched-budget frontier gain defined in Eq.~\eqref{eq:app-pf-area}. It uses the same budget of 3 descendants for every generator family. {The reported active generator is selected from the parent and eligible descendants using evolution-pool results under the realism-constrained archive rule, with $R\ge 0.95$ and relative tolerance $\epsilon=0.05$. The active generator and the preselected PF-Area@3 descendants are then frozen before held-out evaluation.} In Table~\ref{tab:metadrive-paper-baseline-profile-detail}, the Endpoint columns report the best evaluated descendant from the local evolution trace, whereas the \method{} columns report the active generator selected by the archive rule. Table~\ref{tab:metadrive-main-generator-evolution} averages over the IDM and RL ego policies, while Appendix~\ref{app:supplementary-results} keeps the policy-specific rows.

\subsection{CARLA/SafeBench Protocol}
\label{app:carla-setup}

\paragraph{{Benchmark and scenario families.}}
Our study in CARLA~\cite{dosovitskiy2017carla} uses SafeBench~\cite{xu2022safebench} as the closed-loop benchmark and follows the eight CARLA Challenge traffic families used by ChatScene~\cite{zhang2024chatscene}: straight obstacle, turning obstacle, lane changing, vehicle passing, red-light running, unprotected left turn, right turn, and crossing negotiation. {These families are initialized in simulation based on the NHTSA pre-crash taxonomy.} Each family is evaluated over a fixed set of 10 routes, and all generator comparisons use the same ego policy (learned RL policy) and benchmark metrics. During generator evolution, the ego policy is also kept fixed, and the outer loop only changes the scenario generator or structured scenario parameters.

\paragraph{Generator baselines and evolution protocol.}
We instantiate four CARLA generator baselines: parameterized scenarios by AdvSim~\cite{wang2021advsim}, adversarial trajectory optimization (AT)~\cite{zhang2022adversarial}, LLM-based promptable scenario generator by ChatScene~\cite{zhang2024chatscene}, and a human-designed structured scenario generator (Human). For each baseline, the parent set defines the initial scenario archive. Evolution then proceeds for 4 rounds. In each round, candidates are rendered into CARLA-compatible scenarios, checked for simulator compatibility, evaluated by SafeBench~\cite{xu2022safebench}, and inserted into a per-family and per-route archive only from real simulator labels. Final reporting uses a balanced selection rule that keeps the same number of selected scenes per family-route group whenever eligible scenes are available. Therefore, the aggregate results are not dominated by easier families or routes.

\paragraph{{Metrics.}}
We report collision rate (CR), SafeBench overall score (OS), adversarial realism score (ARS), and validity. CR measures how often the fixed ego policy collides under a generated scenario, so a higher CR indicates a stronger attack. OS is the SafeBench driving score~\cite{xu2022safebench}, where lower OS indicates worse ego performance and therefore a more challenging adversarial scene. ARS summarizes the realism diagnostics of the adversarial actors. For a generated candidate $c$, we compute
\begin{equation}
    \mathrm{ARS}(c)
    =
    \mathrm{ValidRate}(c)
    \left(
    0.40\,S_{\mathrm{kin}}(c)
    +0.30\,S_{\mathrm{map}}(c)
    +0.20\,S_{\mathrm{int}}(c)
    +0.10\,S_{\mathrm{dist}}(c)
    \right),
    \label{eq:app-carla-ars}
\end{equation}
where $\mathrm{ValidRate}$ is the fraction of evaluated scenes that pass the scenario validity checks. $S_{\mathrm{kin}}$ measures whether adversarial actors follow bounded and smooth motion, $S_{\mathrm{map}}$ measures consistency with the road map and route geometry, $S_{\mathrm{int}}$ measures whether the adversarial actors maintain plausible interaction patterns with the ego vehicle and other traffic participants, and $S_{\mathrm{dist}}$ measures distance consistency, including non-overlap and reasonable spatial separation. Higher ARS thus indicates a more realistic scene. To report attack-realism trade-offs in Table~\ref{tab:carla-main-generator-evolution}, we define
\begin{equation}
    A_{\mathrm{CARLA}}(c)=\mathrm{CR}(c)\bigl(1-\mathrm{OS}(c)\bigr),
    \qquad
    R_{\mathrm{CARLA}}(c)=\mathrm{ARS}(c),
    \label{eq:app-carla-objectives}
\end{equation}
For evolution traces in Figures~\ref{fig:case-evolution-carla-chatscene}-\ref{fig:case-evolution-carla-human}, we use
\begin{equation}
    S_{\mathrm{trace}}(c)=100\cdot A_{\mathrm{CARLA}}(c)\cdot R_{\mathrm{CARLA}}(c)\cdot \mathrm{Valid}(c)
    \label{eq:app-carla-trace-score}
\end{equation}
only as a compact visualization score for best-so-far trajectory progress. Table and frontier reports still keep attack and realism as separate objectives.

For PF-Area@3, each method is decomposed into family-route groups. Within each group, the parent point is compared with up to 3 eligible evolved descendants under a fixed order, and the hypervolume gain is computed in the Attack-ARS plane following Eq.~\eqref{eq:app-pf-area}. Groups without an eligible evolved descendant contribute zero gain, and the method-level value is the arithmetic mean across groups.

\subsection{Diagnostic and Ablation Studies}
\label{app:diagnostic-setup}

\paragraph{Edit profile analysis.}
Figure~\ref{fig:patch-surface-distribution} counts non-exclusive edit categories over evaluated candidate lineages. A single candidate may contribute to multiple categories when its accepted edit changes both the operator graph and a mechanism module. This diagnostic characterizes the kinds of changes made by the agents during the evolution process.

\paragraph{Actor-critic diagnostics.}
Table~\ref{tab:actor-critic-diagnostic} reports how often critic agents review interventions before the real simulator evaluation. A reviewed proposal is counted as intervened when at least one critic requests a repair or raises a blocking issue. A repaired proposal is counted when the actor returns a revised edit after the critic's feedback. A rejected proposal is counted when the critic cascade or deterministic validator stops the candidate before the simulator evaluation.

\paragraph{Search-width diagnostic.}
Table~\ref{tab:kwidth-sensitivity} compares a narrow single-candidate setting with the parallel proposal width used in MetaDrive. The reported quantities are generated candidates, real-evaluated candidates, validated descendants, and wall-clock speedup. The diagnostic measures search yield and proposal breadth under fixed evolution mechanics, using the same validation, routing, and archive update rules as the main runs.

\paragraph{Attack-only and scalar-search baselines.}
Table~\ref{tab:attack-only-openevolve-ablation} compares the full multi-objective loop with an attack-only variant and an OpenEvolve-based \cite{sharma2025openevolve} scalar-search variant on the same setting. All rows are evaluated with the same attack and realism metrics. The comparison isolates the effect of keeping Pareto archive pressure and realism-aware acceptance during generator evolution.

\paragraph{World-guided candidate routing.}
Figure~\ref{fig:world-sidecar} studies candidate routing before expensive simulator rollout. Candidate pools with real labels are ordered either randomly or by the world evaluator. The figure reports the number of real evaluations required to reach high-quality candidates, the top-$k$ yield, and pairwise ranking quality. The world evaluator is used for routing and prioritization.

\subsection{Downstream Policy Training Validation}
\label{app:sac-setup}

\paragraph{MetaDrive downstream RL fine-tuning.}
This downstream study evaluates whether evolved scenarios remain useful after generator evolution has stopped. We train SAC policies under the scenarios generated by the original ADV-BMT and the evolved ADV-BMT with three random runs. In all cases, the evolved scenario sources are fixed before SAC training begins, and no policy-training signal is fed back into the generator archive. {We use 400 scenarios for RL training, combining the 240 evolution scenes with 160 additional scenes.}

\paragraph{SafeBench downstream RL fine-tuning.}
This study follows the same separation between generator evolution and policy fine-tuning. We follow the same RL fine-tuning protocol as in \cite{zhang2024chatscene}. 
We first freeze the original and the evolved scenario sets from ChatScene, and then fine-tune SAC policies on the corresponding training scenarios. The fine-tuned policies are evaluated on route-disjoint test scenarios, with CR and OS reported in Table~\ref{tab:carla-chatscene-finetune}.

\paragraph{Evaluation metrics.}
For MetaDrive, trained SAC policies are evaluated on frozen 100 adversarial scenarios. Figure~\ref{fig:sac-adv-bmt} reports adversarial crash rate and route completion over training. Lower crash rate and higher route completion indicate better downstream robustness. For SafeBench, Table~\ref{tab:carla-chatscene-finetune} reports CR and OS on the route-disjoint test set. 

\subsection{{LLM Configuration and Computational Cost}}
\label{app:llm-config-cost}

{All LLM agent roles use GPT-5.5 via Codex CLI with \texttt{xhigh} reasoning and otherwise default settings. A representative four-round CARLA run makes 232 role calls, uses 5.097M cached input tokens and 0.159M output tokens, takes 13.5 hours end to end, and has an estimated API cost of US\$5.88.}

% \newpage
\section{Supplementary Results}
\label{app:supplementary-results}

This section reports supplementary results that support the main experimental section.
The results address three questions: whether the evolution loop and its search controls contribute to the reported gains, how the evolution gains vary across ego policies, and how the results change within each scenario family.
All supplementary tables use the same metric directions as the main text: higher Attack, CR, ARS, Realism, and PF-Area are better, while lower RP and OS are better.

\subsection{MetaDrive Ablations and Diagnostics}
\label{app:supp-metadrive-diagnostics}

{Component ablations are reported in Table~\ref{tab:component-ablations} in the main text.}

\paragraph{Multi-objective search.}
Table~\ref{tab:attack-only-openevolve-ablation} compares the full loop with two simpler search variants on the same setting.
Removing the realism-aware multi-objective loop substantially lowers attack while also reducing realism, indicating that both objectives are needed during generator evolution. {The scalar-search method gives the same comparison when candidates are ranked by one combined score rather than by the Pareto archive.}

\begin{table}[!htbp]
\centering
\setlength{\tabcolsep}{4pt}
\small
\begin{threeparttable}
\caption{{Controlled ablation on the same evaluation setting. Attack and Realism are better when higher.}}
\label{tab:attack-only-openevolve-ablation}
\begin{tabular}{lrrrr}
\toprule
Method & Attack & Realism & $\Delta$ Attack & $\Delta$ Realism \\
\midrule
Full \method{} loop & \cellcolor{green!12}0.400 & \cellcolor{green!12}0.983 & +0.0\% & +0.0\% \\
Attack-only search & 0.210 & 0.933 & -47.5\% & -5.0\% \\
OpenEvolve~\cite{sharma2025openevolve} & 0.180 & 0.933 & -55.0\% & -5.0\% \\
\bottomrule
\end{tabular}
\end{threeparttable}
\end{table}

\paragraph{Critic intervention.}
Table~\ref{tab:actor-critic-diagnostic} summarizes how often the critic intervenes before simulator evaluation.
The repair rates show that many proposals are revised before reaching the simulator, while the rejection rates show that the critic still blocks a meaningful fraction of candidates.

\begin{table}[!htbp]
\centering
\setlength{\tabcolsep}{4pt}
\small
\begin{threeparttable}
\caption{{Critic intervention before simulator evaluation. Reviewed is the number of proposals that enter critic review; intervened, repaired, and rejected summarize the pre-simulation filtering process.}}
\label{tab:actor-critic-diagnostic}
\begin{tabular}{lrrrr}
\toprule
Family & Reviewed & Intervened & Repaired & Rejected \\
\midrule
CAT & 200 & 178 (89.0\%) & 177 (88.5\%) & 47 (23.5\%) \\
ADV-BMT & 220 & 192 (87.3\%) & 189 (85.9\%) & 29 (13.2\%) \\
\bottomrule
\end{tabular}
\end{threeparttable}
\end{table}

\paragraph{Proposal width.}
Table~\ref{tab:kwidth-sensitivity} reports candidate yield under different proposal widths.
A narrow single-candidate setting produces no simulator-evaluated descendants, whereas width 20 produces selected and validated candidates across both generator families and ego policies while retaining nearly fourfold wall-clock speedup.

\begin{table*}[!htbp]
\centering
\setlength{\tabcolsep}{4pt}
\small
\begin{threeparttable}
\caption{{Search-width candidate yield. $K$ is the number of parallel proposal slots per epoch; speedup is measured relative to serial proposal execution under the same evaluation protocol.}}
\label{tab:kwidth-sensitivity}
\begin{tabular}{lrrrrr}
\toprule
Setting & K & Proposed & Simulator-eval. & Validated & Speedup \\
\midrule
CAT + IDM (narrow) & 1 & 10 & 0 & 0 & -- \\
CAT + IDM & 20 & 200 & 10 & 8 & 3.83 \\
ADV-BMT + IDM & 20 & 200 & 10 & 9 & 3.92 \\
CAT + RL policy & 20 & 200 & 10 & 6 & 3.84 \\
ADV-BMT + RL policy & 20 & 200 & 10 & 9 & 3.82 \\
\bottomrule
\end{tabular}
\end{threeparttable}
\end{table*}

\subsection{Per-Policy Results in MetaDrive}
\label{app:supp-metadrive-policy}

\paragraph{Policy-specific breakdown.}
Table~\ref{tab:metadrive-paper-baseline-profile-detail} expands the main result in Table~\ref{tab:metadrive-main-generator-evolution} by reporting each generator and ego policy pair separately.
The averaged trends in Table~\ref{tab:metadrive-main-generator-evolution} are consistent with the per-policy view: \method{} improves frontier area across the reported families, while the active generator remains selected from a realism-constrained archive.
This policy-specific view also shows that the same generator family can respond differently to the IDM and RL policies. Rows with zero PF-Area gain indicate that the archive retained the parent or another eligible point under the realism-constrained selection rule.

\begin{table*}[!htbp]
\centering
\scriptsize
\setlength{\tabcolsep}{3pt}
\renewcommand{\arraystretch}{1.1}
\caption{Policy-specific results in MetaDrive. Attack is target-failure rate; RP is $100(1-R)$; Endpoint reports the best evaluated descendant in the local trace; PF-Area@3 is the frontier gain.}
\label{tab:metadrive-paper-baseline-profile-detail}
\resizebox{0.85\linewidth}{!}{
\begin{tabular}{llrrrrrrr}
\toprule
 & & \multicolumn{2}{c}{Parent} & \multicolumn{2}{c}{w/ \method} & \multicolumn{2}{c}{Endpoint} & Frontier \\
\cmidrule(lr){3-4}\cmidrule(lr){5-6}\cmidrule(lr){7-8}\cmidrule(l){9-9}
Method & Policy & Attack$\uparrow$ & RP$\downarrow$ & Attack$\uparrow$ & RP$\downarrow$ & Attack$\uparrow$ & RP$\downarrow$ & PF-Area@3$\uparrow$ \\
\midrule
CAT & IDM & 0.390 & 10.65\% & 0.370 & \cellcolor{green!10}3.84\% & 0.370 & 3.84\% & \cellcolor{green!12}+7.2\% \\
CAT & RL & 0.300 & 10.67\% & 0.290 & \cellcolor{green!10}3.87\% & 0.290 & 3.87\% & \cellcolor{green!12}+7.6\% \\
ADV-BMT & IDM & 0.340 & 2.74\% & \cellcolor{green!10}0.350 & \cellcolor{green!10}1.79\% & 0.350 & 1.79\% & \cellcolor{green!12}+4.0\% \\
ADV-BMT & RL & 0.180 & 1.81\% & \cellcolor{green!10}0.240 & \cellcolor{green!10}1.74\% & 0.240 & 1.74\% & \cellcolor{green!12}+33.4\% \\
AT & IDM & 0.222 & 0.09\% & \cellcolor{green!10}0.280 & 1.50\% & 0.280 & 1.50\% & \cellcolor{green!12}+25.6\% \\
AT & RL & 0.240 & 0.09\% & \cellcolor{green!10}0.250 & 0.62\% & 0.250 & 0.62\% & \cellcolor{green!12}+4.1\% \\
SAGE (w=0.00) & IDM & 0.232 & 0.09\% & \cellcolor{green!10}0.270 & 1.51\% & 0.270 & 1.51\% & \cellcolor{green!12}+16.0\% \\
SAGE (w=0.50) & IDM & 0.232 & 0.06\% & 0.232 & 0.06\% & 0.270 & 1.39\% & \cellcolor{green!12}+16.0\% \\
SAGE (w=1.00) & IDM & 0.232 & 0.04\% & \cellcolor{green!10}0.280 & 1.48\% & 0.280 & 1.48\% & \cellcolor{green!12}+20.2\% \\
SAGE (w=0.00) & RL & 0.180 & 0.09\% & \cellcolor{green!10}0.210 & 0.62\% & 0.210 & 0.62\% & \cellcolor{green!12}+16.6\% \\
SAGE (w=0.50) & RL & 0.200 & 0.06\% & 0.200 & 0.06\% & 0.190 & 0.52\% & +0.0\% \\
SAGE (w=1.00) & RL & 0.190 & 0.04\% & \cellcolor{green!10}0.210 & 0.59\% & 0.210 & 0.59\% & \cellcolor{green!12}+10.5\% \\
LLM-Attacker & IDM & 0.253 & 0.26\% & 0.253 & 0.26\% & 0.250 & 1.48\% & +0.0\% \\
LLM-Attacker & RL & 0.190 & 0.30\% & \cellcolor{green!10}0.210 & 0.76\% & 0.210 & 0.76\% & \cellcolor{green!12}+10.5\% \\
KING & IDM & 0.263 & 0.15\% & \cellcolor{green!10}0.300 & 1.38\% & 0.300 & 1.38\% & \cellcolor{green!12}+14.1\% \\
KING & RL & 0.260 & 0.18\% & 0.260 & 0.18\% & 0.260 & 0.48\% & +0.0\% \\
SEAL & IDM & 0.232 & 0.07\% & \cellcolor{green!10}0.270 & 1.46\% & 0.270 & 1.46\% & \cellcolor{green!12}+16.0\% \\
SEAL & RL & 0.190 & 0.08\% & \cellcolor{green!10}0.220 & 0.59\% & 0.220 & 0.59\% & \cellcolor{green!12}+15.7\% \\
STRIVE & IDM & 0.232 & 0.04\% & \cellcolor{green!10}0.270 & 1.44\% & 0.270 & 1.44\% & \cellcolor{green!12}+16.0\% \\
STRIVE & RL & 0.220 & 0.04\% & 0.220 & 0.04\% & 0.190 & 0.53\% & +0.0\% \\
\bottomrule
\end{tabular}
}
\end{table*}

\subsection{CARLA/SafeBench Detailed Evolution Results}
\label{app:supp-carla-family}

\paragraph{{Family-level progression.}}
{Tables~\ref{tab:carla-as-family-rounds}-\ref{tab:carla-human-family-rounds} expand Table~\ref{tab:carla-main-generator-evolution} by reporting CR, OS, and ARS for every scenario family over all five archive states. The detailed result shows that the aggregate gains are not driven by a single family. AS and AT mainly increase the attack rate while preserving enough realism for frontier growth. ChatScene improves both attack and realism across most families, and the human-designed generator starts with stronger realism but still gains additional ARS through evolution.}

\begin{table*}[!htbp]
\centering
\scriptsize
\setlength{\tabcolsep}{2.pt}
\renewcommand{\arraystretch}{1.1}
\caption{{AdvSim (AS) family-level results by evolution round in CARLA. CR and ARS are better when higher; OS is better when lower.}}
\label{tab:carla-as-family-rounds}
\resizebox{0.98\linewidth}{!}{
\begin{tabular}{llccccccccc}
\toprule
Metric & Round & Straight Obs. & Turn. Obs. & Lane Change & Veh. Pass. & Red-light & Unprot. Left & Right-turn & Crossing & Avg. \\
\midrule
CR$\uparrow$ & 0 & 0.609 & 0.400 & 0.200 & 0.900 & 0.650 & 0.830 & 0.700 & 0.760 & 0.631 \\
 & 1 & 0.597 & 0.450 & 0.158 & 0.975 & 0.651 & 0.856 & 0.698 & 0.854 & 0.655 \\
 & 2 & 0.670 & 0.520 & 0.190 & \textbf{0.980} & 0.940 & \textbf{0.960} & 0.748 & 0.980 & 0.748 \\
 & 3 & 0.690 & 0.580 & \textbf{0.210} & \textbf{0.980} & \textbf{1.000} & \textbf{0.960} & 0.830 & \textbf{1.000} & 0.781 \\
 & 4 & \textbf{0.710} & \textbf{0.610} & \textbf{0.210} & \textbf{0.980} & 0.990 & \textbf{0.960} & \textbf{0.870} & \textbf{1.000} & \cellcolor{green!12}\textbf{0.791} \\
\midrule
OS$\downarrow$ & 0 & 0.607 & 0.688 & 0.793 & 0.457 & 0.625 & 0.540 & 0.505 & 0.494 & 0.589 \\
 & 1 & 0.638 & 0.663 & 0.805 & \textbf{0.436} & 0.634 & 0.534 & 0.501 & 0.454 & 0.583 \\
 & 2 & 0.609 & 0.627 & 0.790 & \textbf{0.436} & 0.490 & 0.485 & 0.481 & 0.389 & 0.538 \\
 & 3 & 0.597 & 0.598 & \textbf{0.783} & \textbf{0.436} & \textbf{0.458} & 0.484 & 0.441 & \textbf{0.378} & 0.522 \\
 & 4 & \textbf{0.588} & \textbf{0.583} & \textbf{0.783} & \textbf{0.436} & 0.463 & \textbf{0.483} & \textbf{0.421} & \textbf{0.378} & \cellcolor{green!12}\textbf{0.517} \\
\midrule
ARS$\uparrow$ & 0 & 0.664 & 0.000 & \textbf{0.562} & 0.998 & \textbf{0.029} & 0.888 & 0.653 & 0.561 & 0.544 \\
 & 1 & 0.891 & 0.000 & 0.512 & 0.998 & 0.014 & 0.969 & 0.743 & \textbf{0.606} & 0.592 \\
 & 2 & 0.980 & 0.000 & 0.529 & 0.998 & 0.010 & \textbf{0.978} & 0.760 & 0.585 & 0.605 \\
 & 3 & 0.980 & 0.000 & 0.550 & 0.998 & 0.010 & \textbf{0.978} & 0.794 & 0.586 & 0.612 \\
 & 4 & \textbf{0.981} & 0.000 & 0.550 & 0.998 & 0.019 & \textbf{0.978} & \textbf{0.804} & 0.586 & \cellcolor{green!12}\textbf{0.615} \\
\bottomrule
\end{tabular}
}
\end{table*}

\begin{table*}[!htbp]
\centering
\scriptsize
\setlength{\tabcolsep}{2.pt}
\renewcommand{\arraystretch}{1.1}
\caption{{Adversarial-trajectory (AT) family-level results by evolution round in CARLA. CR and ARS are better when higher; OS is better when lower.}}
\label{tab:carla-at-family-rounds}
\resizebox{0.98\linewidth}{!}{
\begin{tabular}{llccccccccc}
\toprule
Metric & Round & Straight Obs. & Turn. Obs. & Lane Change & Veh. Pass. & Red-light & Unprot. Left & Right-turn & Crossing & Avg. \\
\midrule
CR$\uparrow$ & 0 & \textbf{0.553} & 0.300 & 0.380 & 0.900 & 0.900 & 0.680 & 0.140 & 0.440 & 0.537 \\
 & 1 & 0.523 & 0.307 & 0.452 & 0.901 & 0.881 & 0.748 & 0.093 & 0.528 & 0.554 \\
 & 2 & 0.540 & 0.460 & 0.630 & 0.900 & 0.940 & 0.880 & 0.150 & 0.700 & 0.650 \\
 & 3 & 0.530 & 0.530 & 0.670 & \textbf{0.910} & \textbf{1.000} & 0.940 & 0.190 & 0.760 & 0.691 \\
 & 4 & 0.530 & \textbf{0.570} & \textbf{0.810} & \textbf{0.910} & \textbf{1.000} & \textbf{0.950} & \textbf{0.250} & \textbf{0.810} & \cellcolor{green!12}\textbf{0.729} \\
\midrule
OS$\downarrow$ & 0 & \textbf{0.631} & 0.737 & 0.684 & \textbf{0.457} & 0.496 & 0.613 & 0.789 & 0.660 & 0.633 \\
 & 1 & 0.662 & 0.729 & 0.656 & 0.473 & 0.515 & 0.584 & 0.800 & 0.620 & 0.630 \\
 & 2 & 0.660 & 0.653 & 0.584 & 0.477 & 0.487 & 0.518 & 0.771 & 0.532 & 0.585 \\
 & 3 & 0.662 & 0.618 & 0.567 & 0.472 & \textbf{0.456} & 0.490 & 0.751 & 0.501 & 0.565 \\
 & 4 & 0.657 & \textbf{0.598} & \textbf{0.516} & 0.472 & \textbf{0.456} & \textbf{0.485} & \textbf{0.721} & \textbf{0.475} & \cellcolor{green!12}\textbf{0.548} \\
\midrule
ARS$\uparrow$ & 0 & \textbf{0.486} & 0.000 & 0.649 & \textbf{0.983} & 0.086 & 0.933 & 0.543 & 0.434 & 0.514 \\
 & 1 & 0.429 & 0.000 & 0.665 & 0.971 & 0.076 & 0.940 & 0.556 & 0.485 & 0.515 \\
 & 2 & 0.412 & 0.000 & 0.761 & 0.978 & 0.117 & \textbf{0.976} & 0.635 & 0.652 & 0.566 \\
 & 3 & 0.413 & 0.000 & 0.781 & 0.978 & 0.146 & 0.975 & 0.694 & 0.740 & 0.591 \\
 & 4 & 0.413 & 0.000 & \textbf{0.846} & 0.974 & \textbf{0.185} & \textbf{0.976} & \textbf{0.714} & \textbf{0.750} & \cellcolor{green!12}\textbf{0.607} \\
\bottomrule
\end{tabular}
}
\end{table*}

\begin{table*}[!htbp]
\centering
\scriptsize
\setlength{\tabcolsep}{2.pt}
\renewcommand{\arraystretch}{1.1}
\caption{{ChatScene family-level results by evolution round in CARLA. CR and ARS are better when higher; OS is better when lower.}}
\label{tab:carla-chatscene-family-rounds}
\resizebox{0.98\linewidth}{!}{
\begin{tabular}{llccccccccc}
\toprule
Metric & Round & Straight Obs. & Turn. Obs. & Lane Change & Veh. Pass. & Red-light & Unprot. Left & Right-turn & Crossing & Avg. \\
\midrule
CR$\uparrow$ & 0 & 0.930 & 0.770 & 0.510 & 0.970 & 0.820 & 0.750 & 0.820 & \textbf{0.930} & 0.812 \\
 & 1 & 0.910 & 0.750 & 0.750 & 0.970 & 0.950 & \textbf{0.890} & 0.950 & 0.910 & 0.885 \\
 & 2 & 0.960 & 0.790 & 0.830 & 0.980 & 0.990 & \textbf{0.890} & \textbf{0.980} & 0.900 & 0.915 \\
 & 3 & 0.980 & 0.840 & \textbf{0.900} & \textbf{1.000} & \textbf{1.000} & \textbf{0.890} & 0.970 & 0.850 & 0.929 \\
 & 4 & \textbf{0.990} & \textbf{0.860} & \textbf{0.900} & \textbf{1.000} & 0.990 & \textbf{0.890} & 0.970 & 0.870 & \cellcolor{green!12}\textbf{0.934} \\
\midrule
OS$\downarrow$ & 0 & 0.479 & 0.512 & 0.680 & 0.432 & 0.548 & 0.590 & 0.469 & \textbf{0.415} & 0.516 \\
 & 1 & 0.488 & 0.533 & 0.564 & 0.412 & 0.482 & 0.521 & 0.430 & 0.426 & 0.482 \\
 & 2 & 0.462 & 0.514 & 0.529 & 0.405 & 0.461 & 0.521 & \textbf{0.412} & 0.431 & 0.467 \\
 & 3 & 0.448 & 0.489 & \textbf{0.494} & 0.397 & \textbf{0.456} & \textbf{0.520} & 0.416 & 0.456 & 0.460 \\
 & 4 & \textbf{0.442} & \textbf{0.479} & \textbf{0.494} & \textbf{0.396} & 0.460 & \textbf{0.520} & 0.416 & 0.446 & \cellcolor{green!12}\textbf{0.457} \\
\midrule
ARS$\uparrow$ & 0 & 0.921 & 0.462 & 0.711 & \textbf{0.978} & 0.464 & 0.573 & 0.764 & 0.960 & 0.729 \\
 & 1 & 0.970 & 0.701 & 0.894 & 0.958 & 0.584 & 0.822 & 0.815 & 0.932 & 0.835 \\
 & 2 & 0.982 & 0.761 & 0.926 & 0.958 & 0.664 & 0.913 & 0.895 & 0.973 & 0.884 \\
 & 3 & 0.985 & \textbf{0.811} & \textbf{0.967} & 0.968 & 0.773 & 0.944 & 0.906 & \textbf{0.984} & 0.917 \\
 & 4 & \textbf{0.987} & 0.792 & 0.957 & 0.968 & \textbf{0.823} & \textbf{0.955} & \textbf{0.926} & \textbf{0.984} & \cellcolor{green!12}\textbf{0.924} \\
\bottomrule
\end{tabular}
}
\end{table*}

\begin{table*}[!htbp]
\centering
\scriptsize
\setlength{\tabcolsep}{2.pt}
\renewcommand{\arraystretch}{1.1}
\caption{{Human-designed generator family-level results by evolution round in CARLA. CR and ARS are better when higher; OS is better when lower.}}
\label{tab:carla-human-family-rounds}
\resizebox{0.98\linewidth}{!}{
\begin{tabular}{llccccccccc}
\toprule
Metric & Round & Straight Obs. & Turn. Obs. & Lane Change & Veh. Pass. & Red-light & Unprot. Left & Right-turn & Crossing & Avg. \\
\midrule
CR$\uparrow$ & 0 & \textbf{0.967} & \textbf{0.689} & 0.788 & 0.552 & 0.479 & 0.557 & 0.758 & 0.522 & 0.664 \\
 & 1 & 0.947 & 0.665 & 0.810 & \textbf{0.562} & 0.760 & 0.700 & 0.750 & \textbf{0.701} & 0.737 \\
 & 2 & 0.917 & 0.611 & \textbf{0.850} & \textbf{0.562} & \textbf{0.820} & \textbf{0.720} & 0.790 & 0.681 & \cellcolor{green!12}\textbf{0.744} \\
 & 3 & 0.928 & 0.548 & 0.840 & 0.552 & 0.780 & 0.690 & \textbf{0.800} & 0.666 & 0.725 \\
 & 4 & 0.948 & 0.471 & 0.840 & 0.552 & 0.730 & 0.700 & 0.790 & 0.671 & 0.713 \\
\midrule
OS$\downarrow$ & 0 & 0.466 & \textbf{0.607} & 0.556 & 0.652 & 0.684 & 0.630 & 0.565 & 0.657 & 0.602 \\
 & 1 & 0.470 & 0.617 & 0.544 & 0.645 & 0.558 & 0.571 & 0.567 & \textbf{0.571} & 0.568 \\
 & 2 & 0.480 & 0.647 & \textbf{0.522} & \textbf{0.644} & \textbf{0.527} & \textbf{0.562} & 0.547 & 0.582 & \cellcolor{green!12}\textbf{0.564} \\
 & 3 & 0.474 & 0.678 & 0.527 & 0.649 & 0.544 & 0.573 & \textbf{0.542} & 0.588 & 0.572 \\
 & 4 & \textbf{0.464} & 0.716 & 0.527 & 0.649 & 0.568 & 0.568 & 0.547 & 0.585 & 0.578 \\
\midrule
ARS$\uparrow$ & 0 & 0.877 & 0.731 & 0.873 & 0.841 & 0.934 & 0.924 & 0.814 & 0.851 & 0.856 \\
 & 1 & 0.880 & 0.796 & 0.915 & 0.899 & 0.914 & 0.939 & 0.894 & 0.867 & 0.888 \\
 & 2 & 0.910 & \textbf{0.822} & 0.944 & 0.900 & 0.922 & 0.948 & 0.902 & 0.868 & 0.902 \\
 & 3 & \textbf{0.922} & 0.799 & 0.945 & 0.939 & 0.934 & \textbf{0.949} & 0.931 & 0.899 & 0.915 \\
 & 4 & \textbf{0.922} & 0.794 & \textbf{0.965} & \textbf{0.949} & \textbf{0.946} & \textbf{0.949} & \textbf{0.951} & \textbf{0.939} & \cellcolor{green!12}\textbf{0.927} \\
\bottomrule
\end{tabular}
}
\end{table*}

\subsection{{Sensitivity to Generator Initialization}}
\label{app:generator-initialization}

{We sample three parameter starts from the original ranges for each of two ChatScene families: Turning Obstacle and Lane Change. Across all six runs, CR, ARS, and validity increase and OS decreases (Table~\ref{tab:generator-initialization}).}

\begin{table}[!htbp]
\centering
\scriptsize
\setlength{\tabcolsep}{8pt}
\renewcommand{\arraystretch}{1.1}
\caption{{Generator initialization sensitivity across three starts per family. Initial and Final report scores before and after generator evolution.}}
\label{tab:generator-initialization}
\resizebox{0.9\linewidth}{!}{%
{%
\begin{tabular}{lcccc}
\toprule
Metric & \multicolumn{2}{c}{Turning Obstacle} & \multicolumn{2}{c}{Lane Change} \\
\cmidrule(lr){2-3}\cmidrule(lr){4-5}
 & Initial & Final & Initial & Final \\
\midrule
CR$\uparrow$ & $0.6833 \pm 0.0252$ & $\mathbf{0.8933 \pm 0.0896}$ & $0.5833 \pm 0.0321$ & $\mathbf{0.8967 \pm 0.0321}$ \\
OS$\downarrow$ & $0.5546 \pm 0.0122$ & $\mathbf{0.4661 \pm 0.0407}$ & $0.6471 \pm 0.0167$ & $\mathbf{0.4948 \pm 0.0171}$ \\
ARS$\uparrow$ & $0.4818 \pm 0.0201$ & $\mathbf{0.7088 \pm 0.0252}$ & $0.7211 \pm 0.0378$ & $\mathbf{0.9739 \pm 0.0154}$ \\
Validity$\uparrow$ & $0.4900 \pm 0.0200$ & $\mathbf{0.7167 \pm 0.0252}$ & $0.7267 \pm 0.0379$ & $\mathbf{0.9767 \pm 0.0153}$ \\
\bottomrule
\end{tabular}%
}%
}
\end{table}

\subsection{{Cross-Policy Scenario Transfer}}
\label{app:policy-transfer}

{We freeze the original and evolved scenarios selected under SAC and evaluate both with CARLA BasicAgent, a rule-based controller. Table~\ref{tab:policy-transfer} shows higher attack for the evolved scenarios under both SAC and BasicAgent.}

\begin{table}[!htbp]
\centering
\scriptsize
\setlength{\tabcolsep}{8pt}
\renewcommand{\arraystretch}{1.1}
\caption{{Cross-policy evaluation of frozen original and evolved scenarios selected under SAC.}}
\label{tab:policy-transfer}
\resizebox{0.7\linewidth}{!}{%
{%
\begin{tabular}{lcc}
\toprule
Ego policy & Original Attack$\uparrow$ & Evolved Attack$\uparrow$ \\
\midrule
Learned SAC (search policy) & 0.3759 & \textbf{0.4477} \\
CARLA BasicAgent & 0.0612 & \textbf{0.1434} \\
\bottomrule
\end{tabular}%
}%
}
\end{table}

\subsection{Additional Evolution Traces}
\label{app:supp-evolution-traces}

\begin{figure*}[!htbp]
\centering
\includegraphics[width=0.78\linewidth]{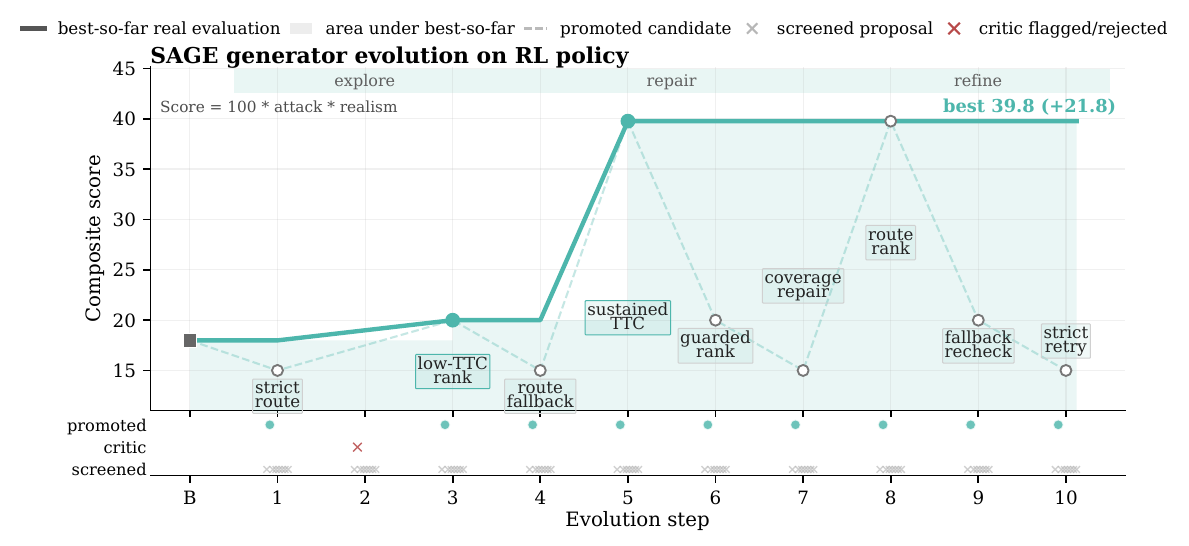}
\caption{{SAGE evolution trajectory on the RL policy. Markers show screened, evaluated-but-not-promoted, and promoted proposals at each local step.}}
\label{fig:case-evolution-sage}
\end{figure*}

\begin{figure*}[!htbp]
\centering
\includegraphics[width=0.78\linewidth]{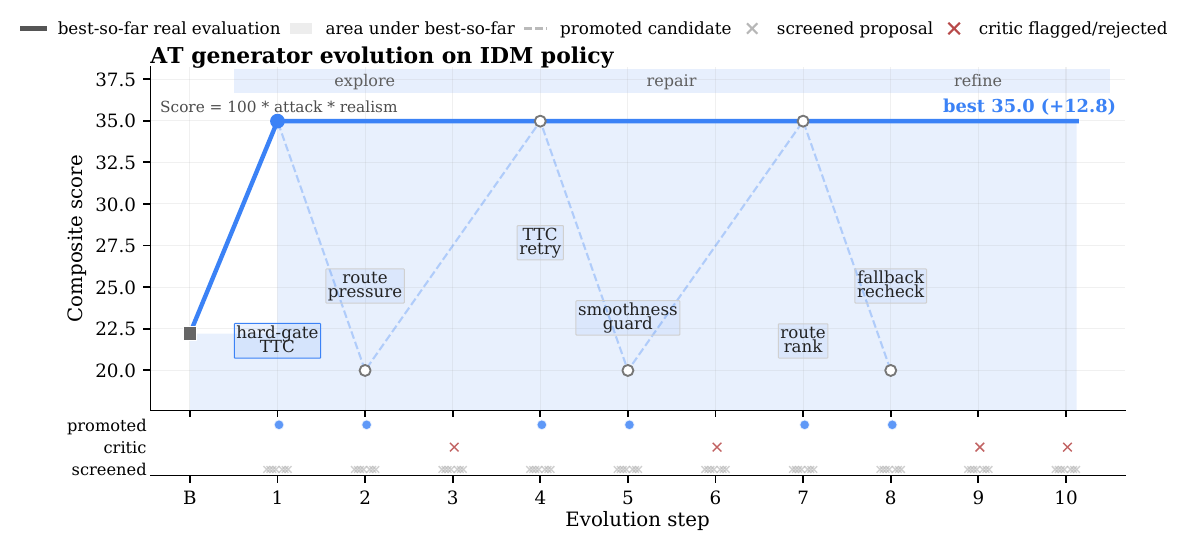}
\caption{{AT evolution trajectory on the IDM policy. Markers show screened, evaluated-but-not-promoted, and promoted proposals at each local step.}}
\label{fig:case-evolution-at}
\end{figure*}

\begin{figure*}[!htbp]
\centering
\includegraphics[width=0.78\linewidth]{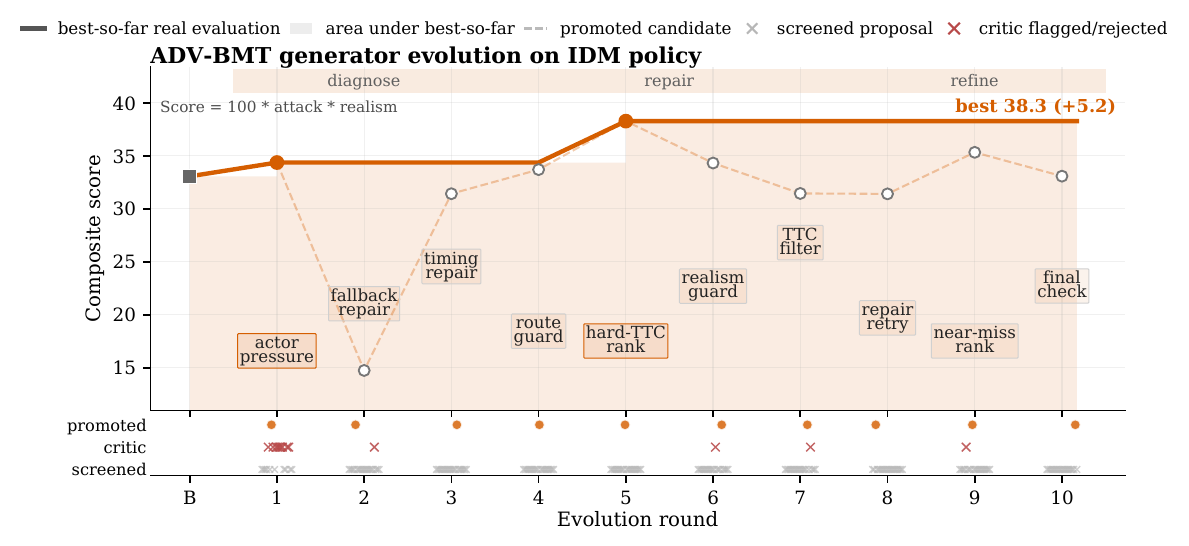}
\caption{{ADV-BMT evolution trajectory on the IDM policy. Markers show screened, evaluated-but-not-promoted, and promoted proposals at each local step.}}
\label{fig:case-evolution-advbmt-idm}
\end{figure*}

\paragraph{MetaDrive.}
Figure~\ref{fig:case-evolution-sage}, Figure~\ref{fig:case-evolution-at}, and Figure~\ref{fig:case-evolution-advbmt-idm} provide additional evolution traces of generators in MetaDrive.
Across these examples, the promoted generators are reached through mechanism-level edits such as actor ranking, route checks, time-to-conflict filters, fallback repair, and realism guards.
Together with Figure~\ref{fig:case-evolution-main}, these traces show that the search process accumulates interpretable generator changes without selecting isolated random perturbations.

\begin{figure*}[!htbp]
\centering
\includegraphics[width=0.78\linewidth]{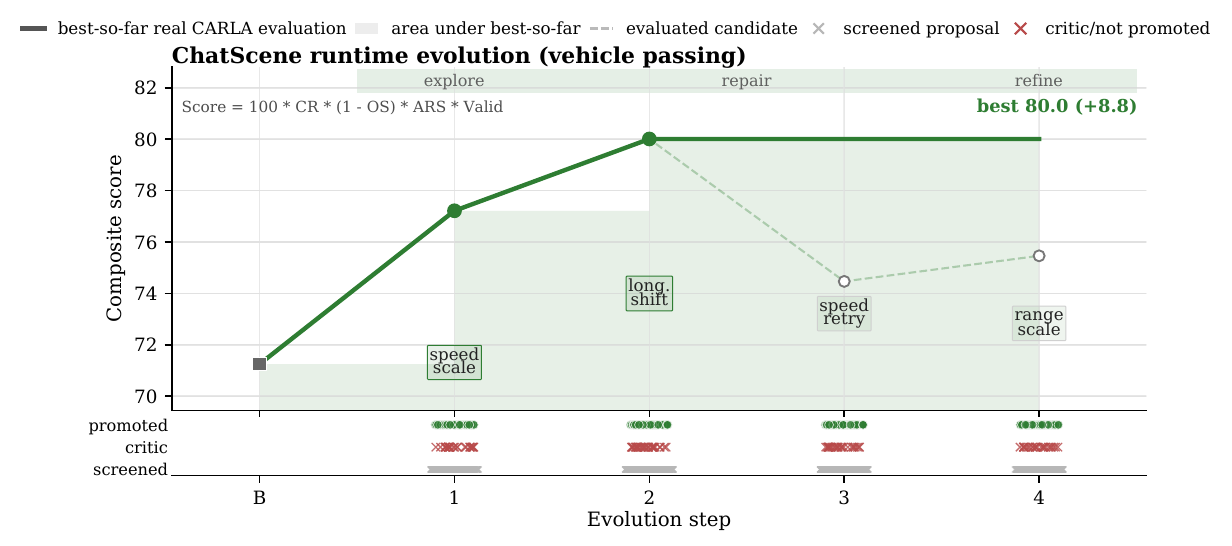}
\caption{ChatScene evolution trace for vehicle-passing scenarios in CARLA. The score is $100\cdot\mathrm{CR}\cdot(1-\mathrm{OS})\cdot\mathrm{ARS}\cdot\mathrm{Valid}$. Markers show screened, evaluated-but-not-promoted, and promoted proposals at each local step.}
\label{fig:case-evolution-carla-chatscene}
\end{figure*}

\begin{figure*}[!htbp]
\centering
\includegraphics[width=0.78\linewidth]{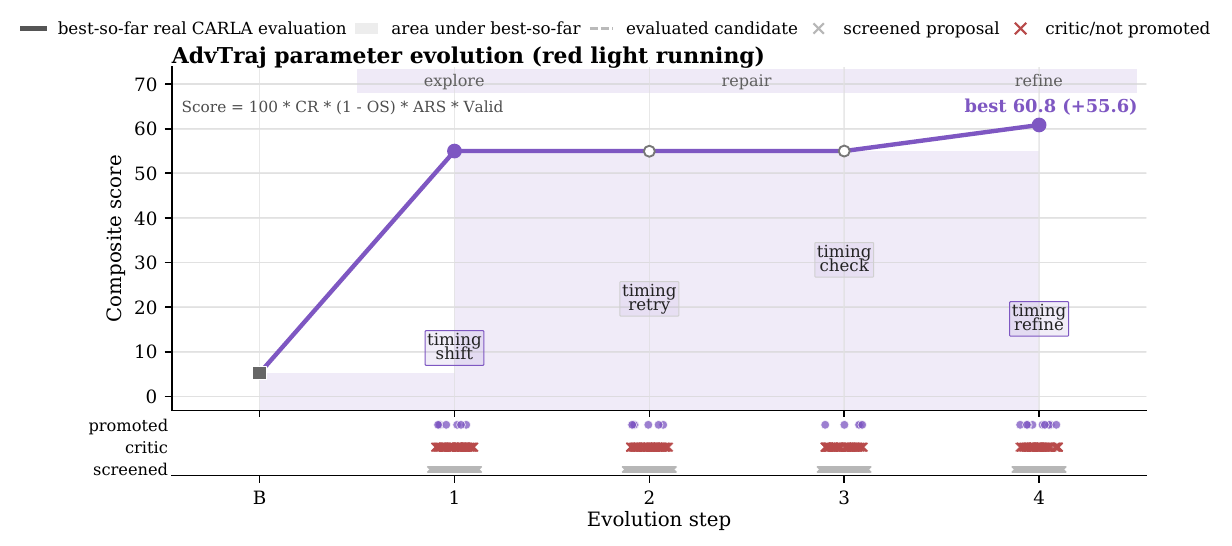}
\caption{AdvTraj evolution trace for red-light-running scenarios in CARLA. The score is $100\cdot\mathrm{CR}\cdot(1-\mathrm{OS})\cdot\mathrm{ARS}\cdot\mathrm{Valid}$. Markers show screened, evaluated-but-not-promoted, and promoted proposals at each local step.}
\label{fig:case-evolution-carla-advtraj}
\end{figure*}

\begin{figure*}[!htbp]
\centering
\includegraphics[width=0.78\linewidth]{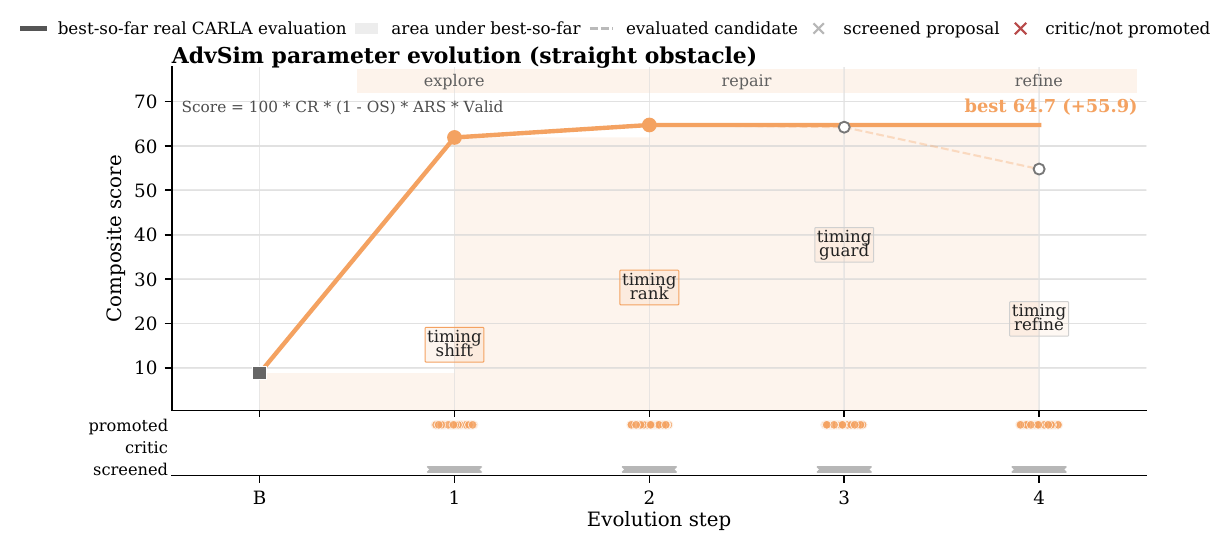}
\caption{AdvSim evolution trace for straight-obstacle scenarios in CARLA. The score is $100\cdot\mathrm{CR}\cdot(1-\mathrm{OS})\cdot\mathrm{ARS}\cdot\mathrm{Valid}$. Markers show screened, evaluated-but-not-promoted, and promoted proposals at each local step.}
\label{fig:case-evolution-carla-advsim}
\end{figure*}

\begin{figure*}[!htbp]
\centering
\includegraphics[width=0.78\linewidth]{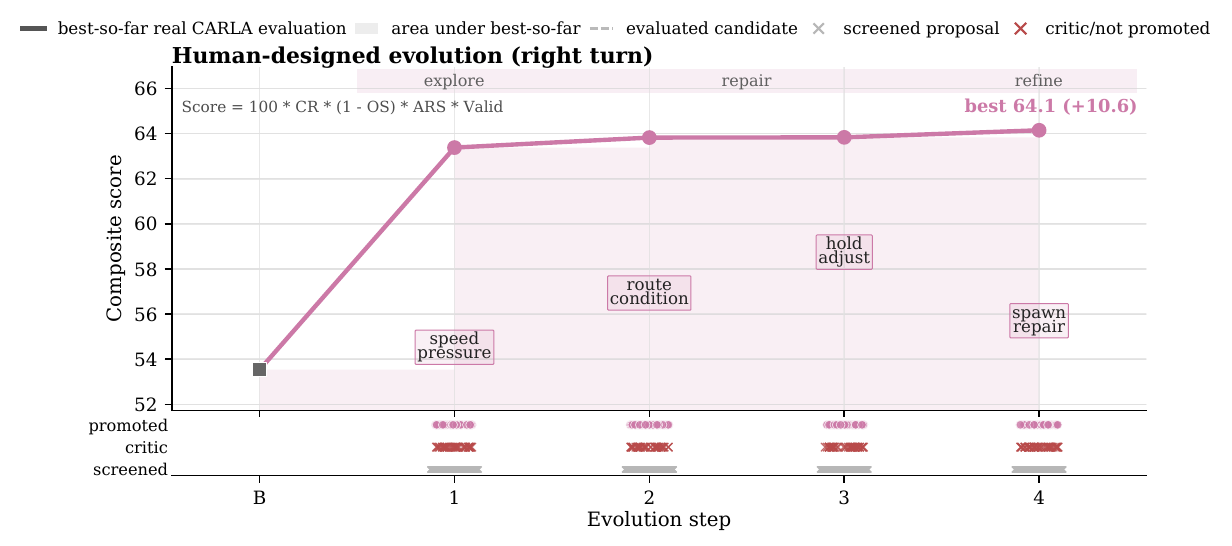}
\caption{Human-designed baseline evolution trace for right-turn scenarios in CARLA. The score is $100\cdot\mathrm{CR}\cdot(1-\mathrm{OS})\cdot\mathrm{ARS}\cdot\mathrm{Valid}$. Markers show screened, evaluated-but-not-promoted, and promoted proposals at each local step.}
\label{fig:case-evolution-carla-human}
\end{figure*}

\paragraph{{CARLA/SafeBench.}}
{Figures~\ref{fig:case-evolution-carla-chatscene}-\ref{fig:case-evolution-carla-human} provide the corresponding evolution traces of the SafeBench baselines. Each plot uses local evolution steps on the x-axis, making the trace comparable within a scenario family. 
% The marker rows distinguish screened proposals, evaluated candidates that were not promoted, and promoted candidates. 
Across the four epochs, the best-so-far score improves over multiple evaluated candidates instead of at a single endpoint.}

% \clearpage
\section{Core Architectural Overview and System Prompts}
\label{app:architecture-prompts}

\subsection{Architectural Overview}
\label{app:architecture-details}

Figure~\ref{fig:system-arch} shows the core implementation organization of \method{}. We display the main modules that correspond to the algorithmic state in Appendix~\ref{app:extended-method}. 

\begin{figure*}[!htbp]
\centering
\begin{tcolorbox}[
title=System Architecture of \method,
width=0.96\textwidth
]
\begin{minted}{markdown}
EvoDrive/
  evodrive/
    agents/
      actor templates        bounded generator-patch proposals
      critic templates       structural, attack, realism, and isolation review
      repair templates       one-step patch repair after critic feedback
      response parser        JSON-only runtime response normalization
    adapters/
      family contracts       editable ranges and operator families
      baseline registry      imported generator-family interfaces
      compile contracts      backend constraints and simulator targets
    candidates/
      candidate schema       normalized generator candidate records
      validation gates       schema, parent, interface, isolation, safety, compile
      provenance records     parent hash, patch hash, route, and evidence links
    lineage/
      solution programs      active parent state and content hashes
      program patches        bounded add, replace, and remove edits
      scheduler              lineage selection and active-child promotion
      isolation checks       same-family continuity and rejected-child quarantine
    loop/
      epoch runner           parent selection, proposal, validation, evaluation
      parallel workers       immutable candidate snapshots
      serial acceptor        deterministic archive, lineage, and memory updates
    evaluation/
      compiler               typed program to simulator-operator composition
      simulator bridge       fixed ego policy closed-loop rollout
      metric reducer         attack, realism, diagnostics, and seed summaries
    archive/
      frontier analyzer      non-dominated attack-realism frontier
      acceptance gates       hypervolume, local-frontier, and parent-tradeoff tests
      active selector        practical generator chosen from archive entries
    memory/
      memory cards           failure memories, success motifs, and critic notes
      retrieval policy       causal lineage, family, mechanism, and scene matching
      reflection summaries   next-step repair, reuse, diversify, or pivot hints
    world/
      feature builder        candidate, policy, scene, and interaction features
      router                 select, reject, or audit before real rollout
      calibrator             simulator-labeled correction statistics
      world agent            evaluator snapshot proposal and bounded updates
      world judge            deterministic promotion of evaluator snapshots
    baselines/
      imported families      AT, SAGE, LLM-Attacker, KING, SEAL, STRIVE, ChatScene, etc.
      adapter seeds          normalized parent generators for evolution
    evidence/
      attempt records        validation, routing, simulation, and archive traces
      paper bundles          table and figure summaries
      audit summaries        lineage continuity and reproducibility checks
\end{minted}
\end{tcolorbox}
\vspace{-8pt}
\caption{Implementation organization of \method{}. Runtime agents propose bounded generator edits, deterministic modules validate and compile candidates, simulator evaluation provides labels, and the archive, lineage, memory, and world components maintain the closed-loop evolution state.}
\label{fig:system-arch}
\end{figure*}

\paragraph{State flow.}
The architecture maintains six recurring objects. A \emph{solution program} is the current generator state for one source family; a \emph{program patch} is a bounded edit against that state; a \emph{candidate record} is the normalized object produced after parsing and validation; an \emph{attempt record} stores routing, validation, and evaluation evidence; an \emph{archive entry} stores the accepted simulator-labeled outcome; and a \emph{memory card} summarizes causal evidence that can be retrieved by later actors and critics. At epoch \(t\), the loop selects a parent solution \(p_t\), instantiates an actor prompt with retrieved memory and family-specific edit surfaces, parses a patch \(\delta_t\), and passes it through deterministic validation. Valid candidates are either evaluated directly or routed through the world evaluator. Simulator-labeled outcomes then update the Pareto archive, active lineage state, and memory cards through a serial acceptor, which prevents parallel proposals from creating inconsistent ancestry.

\paragraph{World evaluator.}
The world evaluator component is implemented as a separate evaluator line. It receives candidate, policy, scene, and interaction features, predicts whether a proposal is worth simulator execution, and exposes its uncertainty to an audit route. Its own updates are proposed by the world agent and promoted only by a deterministic judge using simulator-labeled calibration evidence. This separation keeps generator evolution, simulator labels, and archive acceptance as explicit downstream objects while allowing the prescreening policy to improve over time.

\paragraph{Benchmark adapters.}
The same closed-loop architecture is used for MetaDrive and CARLA benchmarks. Benchmark-specific adapters provide the simulator bridge, generator contracts, and metric reducer, while the proposal agents, validation gates, lineage state, Pareto archive, memory retrieval, world routing, and prompt roles remain shared. This keeps benchmark-specific details outside the agent contract and makes results comparable through the normalized attack-realism archive.

\subsection{Prompt Catalog}
\label{app:prompt-catalog}

At runtime, \method{} uses a collection of role-specific prompt templates. Each template is filled with the selected parent state, the corresponding generator interface, retrieved memory, scene or cluster metadata, and the current search target. The templates below present the core instructions. We omit long structured configurations and environment-specific metadata. The displayed fields specify the input-output contract used by the system: each agent receives an explicit parent state, edits only the permitted typed fields, and returns a structured object that can be parsed and checked by the validator.

\begin{tcolorbox}[
promptbox,
title=Shared Runtime I/O Contract,
colback=promptcontract!3!white,
colframe=promptcontract!80!black,
colbacktitle=promptcontract!82!black
]
{\scriptsize\ttfamily
All runtime roles receive: parent solution metadata, source family, parent content hash, objective target, editable surface, protected components, retrieved memory references, and optional cluster references.\\[2pt]
All runtime roles must return: exactly one JSON object, no explanatory prose outside JSON, stable response\_id, explicit parent identifiers, and either a bounded program patch, a critic review, a repaired patch, or a world evaluator candidate.\\[2pt]
Shared invariants: preserve the source family of the parent; do not edit the ego policy, benchmark scenes, simulator evaluator, real labels, archive acceptor, or other lineages; use only supported mechanism modules and typed simulator operators; cite memory references only when they influence the proposed edit or review.\\[2pt]
Failure handling: malformed JSON, parent-hash mismatch, unsupported edit paths, cross-family mechanisms, missing provenance, or non-compilable operator graphs are treated as validation failures before simulation.
}
\end{tcolorbox}

\bigskip
\begin{tcolorbox}[
promptbox,
title=Generator Actor Prompt,
colback=promptactor!3!white,
colframe=promptactor!75!black,
colbacktitle=promptactor!82!black
]
{\scriptsize\ttfamily
Role: EvoDrive-managed runtime agent; runtime role = baseline\_program\_actor.\\
Workflow: baseline\_program\_actor\_critic.\\[2pt]
Goal: improve the current family-specific generator by proposing a bounded program patch against the provided parent solution. Continue the lineage; do not create a fresh mixed-family generator.\\[2pt]
Context provided at runtime: iteration index, fixed ego policy descriptor, lineage id, source family, parent solution id, parent content hash, parent generation, active objective target, retrieved memory cards, cluster summaries, and normalized parent solution JSON.\\[2pt]
Editable surface: /parameters, /operator\_graph, /scenario\_conditioned\_policy, /mechanism\_modules.\\[2pt]
Patch design rules: preserve the source family; modify only a small number of typed fields; explain the intended behavioral mechanism in the hypothesis; prefer edits supported by retrieved memory; keep the operator graph compilable; maintain route feasibility and interaction plausibility before simulator rollout.\\[2pt]
Quality checks before returning: parent identifiers match the input; every operation has a legal path; each parameter lies inside the provided range; added modules specify their typed inputs and outputs; the patch can be interpreted without natural-language side channels.\\[2pt]
Output schema: \{ "response\_id": "...", "program\_patch": \{ "source\_baseline": "...", "parent\_solution\_id": "...", "parent\_program\_hash": "...", "operations": [...], "hypothesis": "...", "memory\_refs\_used": [...], "cluster\_refs\_used": [...] \} \}.
}
\end{tcolorbox}

\bigskip
\begin{tcolorbox}[
promptbox,
title=Critic Prompt,
colback=promptcritic!3!white,
colframe=promptcritic!75!black,
colbacktitle=promptcritic!82!black
]
{\scriptsize\ttfamily
Role: EvoDrive-managed runtime agent; runtime role = structural\_critic, attack\_critic, realism\_critic, or isolation\_critic.\\
Workflow: baseline\_program\_actor\_critic.\\[2pt]
Task: review the proposed program patch before simulation and decide whether it should be approved, repaired once, or rejected. The critic reviews the patch, not the eventual simulator outcome.\\[2pt]
Inputs: parent solution JSON, program patch JSON, ego policy descriptor, lineage id, source family, parent hash, editable surface, protected components, and role-specific focus.\\[2pt]
Structural focus: verify JSON schema, operation order, edit paths, typed module signatures, parameter ranges, and compiler-facing operator graph consistency.\\
Attack focus: check whether the edit plausibly increases failure exposure of the fixed ego policy through interactions such as cut-in pressure, occlusion, route conflict, or speed mismatch.\\
Realism focus: check map validity, route consistency, non-overlap, acceleration, jerk, time-to-collision plausibility, and controller feasibility.\\
Isolation focus: check source-family continuity, parent-hash consistency, protected-component constraints, and absence of cross-lineage mechanisms.\\[2pt]
Decision policy: approve when all blocking checks pass; request repair when a local edit can resolve the issue; reject when the patch changes protected state, lacks a valid parent, mixes source families, or cannot be mapped to supported simulator operators.\\[2pt]
Output schema: \{ "critic\_review": \{ "role": "...", "decision": "approve|repair|reject", "blockers": [...], "repair\_hints": [...], "risk\_tags": [...], "scope": "pre\_simulation" \} \}.
}
\end{tcolorbox}

\bigskip
\begin{tcolorbox}[
promptbox,
title=Actor Repair Prompt,
colback=promptrepair!4!white,
colframe=promptrepair!78!black,
colbacktitle=promptrepair!85!black
]
{\scriptsize\ttfamily
Role: EvoDrive-managed runtime agent; runtime role = baseline\_program\_actor\_repair.\\[2pt]
Task: repair the previous family-specific patch exactly once using critic feedback. Keep the same parent solution id, parent content hash, source family, objective target, and editable surface.\\[2pt]
Inputs: original patch, critic reviews, structural blockers, role-specific risk tags, retrieved memory cards, optional cluster references, parent solution metadata, and legal edit paths.\\[2pt]
Permitted repair actions: reduce unsupported parameter magnitudes; replace an unsupported module with a supported typed mechanism; add route, collision, or acceleration guards; move an operation to a legal edit path; remove a nonessential operation that caused validation failure; clarify the patch hypothesis when the intended mechanism is underspecified.\\[2pt]
Repair discipline: change the smallest portion of the patch needed to clear the blockers; do not introduce a new behavioral idea unrelated to the original hypothesis; do not weaken benchmark, evaluator, archive, or ego-policy constraints; preserve memory references only if they still justify the revised edit.\\[2pt]
Output schema: \{ "program\_patch": \{ "source\_baseline": "...", "parent\_solution\_id": "...", "parent\_program\_hash": "...", "operations": [...], "hypothesis": "...", "repair\_summary": "...", "memory\_refs\_used": [...] \} \}.
}
\end{tcolorbox}

\bigskip
\begin{tcolorbox}[
promptbox,
title=World Agent Prompt,
colback=promptworld!4!white,
colframe=promptworld!78!black,
colbacktitle=promptworld!84!black
]
{\scriptsize\ttfamily
Role: EvoDrive World Agent.\\[2pt]
Task: improve only the standalone world evaluator line. The world agent proposes an evaluator snapshot; the deterministic world judge decides whether the snapshot is promoted.\\[2pt]
Inputs: active evaluator snapshot, candidate-feature schema, recent prediction-quality report, simulator-labeled calibration evidence, false-rejection audits, uncertainty failures, routing statistics, and evaluator-update surface.\\[2pt]
Feature groups available to the evaluator: generator-family metadata, patch-operation descriptors, route and map descriptors, actor interaction summaries, realism diagnostics, memory-derived mechanism tags, uncertainty estimates, and recent calibration residuals.\\[2pt]
Evaluator update surface: feature selection, score calibration, ranking policy, uncertainty policy, audit-routing thresholds, and abstention behavior.\\[2pt]
Protected components: scenario generators, ego policies, simulator evaluator, real labels, archive acceptor, generator lineage state, and candidate program patches.\\[2pt]
Evaluation discipline: optimize prescreening and ranking of unevaluated candidates without altering how real labels are produced or how accepted archive entries are selected; expose uncertainty when the evaluator is outside its calibrated region.\\[2pt]
Output schema: \{ "world\_candidate": \{ "evaluation\_hypothesis": "...", "expected\_judge\_metric\_improvement": "...", "proposed\_evaluator\_configuration": \{...\}, "calibration\_rationale": "...", "audit\_routing\_policy": "...", "risk\_notes": [...] \} \}.
}
\end{tcolorbox}

\end{document}